\documentclass[conference]{IEEEtran}
\IEEEoverridecommandlockouts
\pdfoutput=1
\usepackage{cite}

\usepackage{amsmath,graphicx}

\usepackage{wrapfig}
\usepackage{caption}
\usepackage{subcaption}
\usepackage{textcomp}
\usepackage[dvipsnames]{xcolor}
\usepackage[inline]{enumitem}
\usepackage{orcidlink}
\usepackage{makecell}
\usepackage{array}
\usepackage{booktabs}
\usepackage{multirow}
\usepackage{algorithm}
\usepackage[noend]{algpseudocode}
\begin{document}
% Title.
% ------
\title{Multi-objective Feature Selection in Remote Health Monitoring Applications}

\author{
\IEEEauthorblockN{Le Ngu Nguyen$^{\star}$ \enspace Constantino Álvarez Casado$^{\star}$ \enspace Manuel Lage Cañellas$^{\star}$ \enspace Anirban Mukherjee$^{\star \dagger}$ \enspace Nhi Nguyen$^{\star}$ \\ Dinesh Babu Jayagopi$^{\dagger}$ \enspace Miguel Bordallo López$^{\star}$}

\IEEEauthorblockA{$^{\star}$ Center for Machine Vision and Signal Analysis, University of Oulu, Finland \\
$^{\dagger}$ International Institute of Information Technology, Bangalore, India}
}

%

%\ninept
%

\maketitle
% \IEEEpeerreviewmaketitle % for ANONYMOUS reviewing
%
\begin{abstract}

% Normal Version (submitted)
Radio frequency (RF) signals have facilitated the development of non-contact human monitoring tasks, such as vital signs measurement, activity recognition, and user identification. In some specific scenarios, an RF signal analysis framework may prioritize the performance of one task over that of others. In response to this requirement, we employ a multi-objective optimization approach inspired by biological principles to select discriminative features that enhance the accuracy of breathing patterns recognition while simultaneously impeding the identification of individual users. This approach is validated using a novel vital signs dataset consisting of 50 subjects engaged in four distinct breathing patterns. Our findings indicate a remarkable result: a substantial divergence in accuracy between breathing recognition and user identification. As a complementary viewpoint, we present a contrariwise result to maximize user identification accuracy and minimize the system's capacity for breathing activity recognition.

\end{abstract}


%%%%%%%%%%%%%%%%%%%%%%%%%%%%%%%%%%%%%%%%%%%%%
%                                           %
%               INTRODUCTION                %
%                                           %
%%%%%%%%%%%%%%%%%%%%%%%%%%%%%%%%%%%%%%%%%%%%%
\section{Introduction}
\label{sec:intro}

Non-contact methods for monitoring vital signs have gained increasing attention in recent years, facilitated by advancements in video analysis and radar signal processing technologies. Remote photoplethysmography (rPPG) serves as a representative example, wherein fluctuations in skin light absorption and reflection are analyzed to quantify parameters such as blood volume pulse and heart rate \cite{Casado2023}. In parallel, radar-based approaches utilize high-frequency signals to detect minute thoracic wall movements, facilitating the measurement of heart rate through ballistocardiographic analysis as well as respiration rate \cite{Eder2023}. These non-intrusive techniques offer the significant advantage of not requiring the user to wear any wearable sensors, thus making them more convenient and less obtrusive. Nevertheless, they also introduce potential issues related to user privacy. For instance, the analysis of facial videos for vital sign extraction has been identified as a potential breach of privacy \cite{Gupta2023}. Furthermore, millimeter-wave (mmWave) radar technology has been shown to identify individuals with high accuracy based on gait patterns, which poses additional privacy concerns \cite{Yang2020}. Recent research has also indicated the feasibility of using these non-contact methods for activity recognition or user identification within a small cohort \cite{Nguyen2022}. This multi-purpose utility underscores the need for a careful balance between convenience and privacy in the deployment of such technologies.

\begin{figure}
    \centering
     \begin{subfigure}[b]{0.48\columnwidth}
         \centering
         \includegraphics[width=\columnwidth]{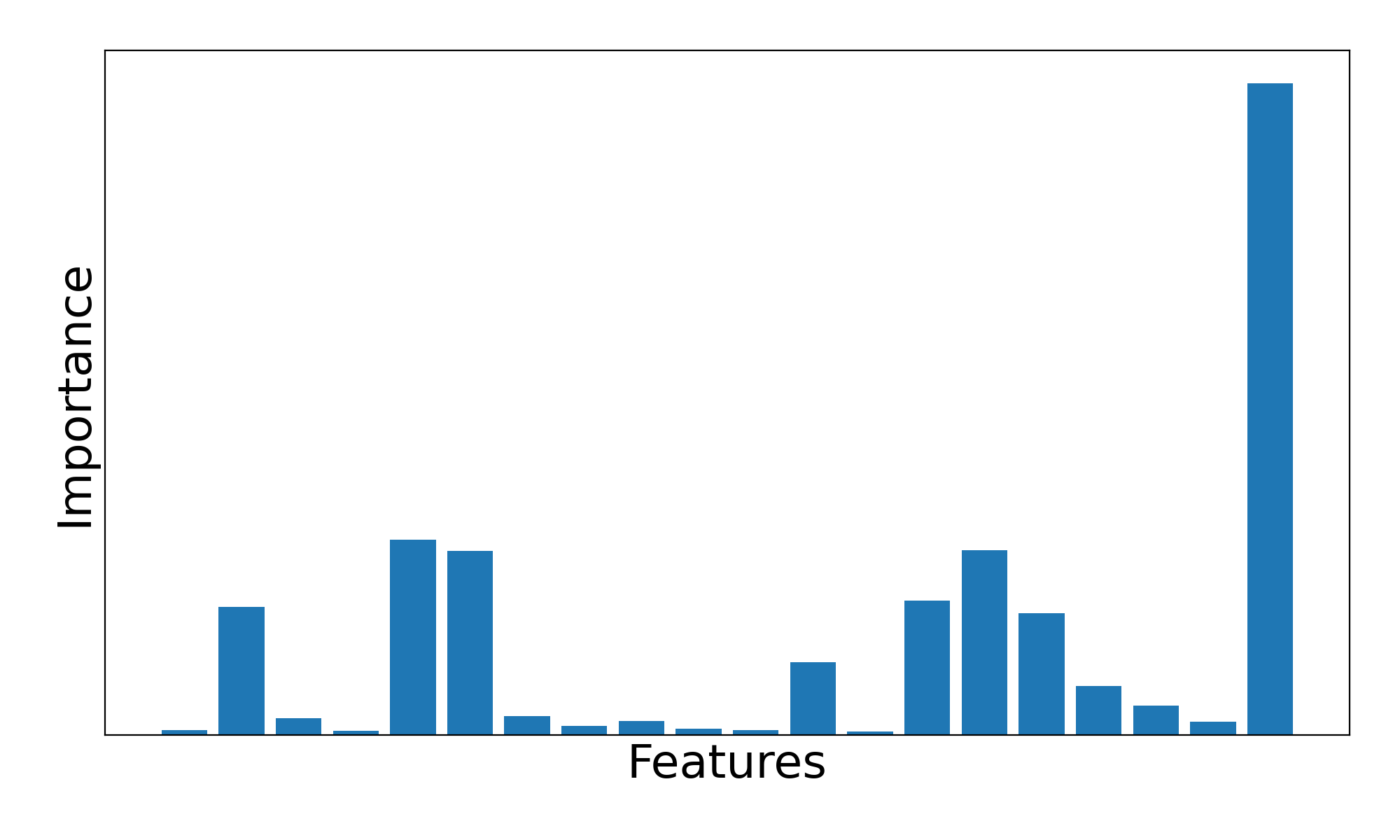}
         \caption{Breathing pattern recognition}
         \label{fig:Recognition_Features}
     \end{subfigure}
     \begin{subfigure}[b]{0.48\columnwidth}
         \centering
         \includegraphics[width=\columnwidth]{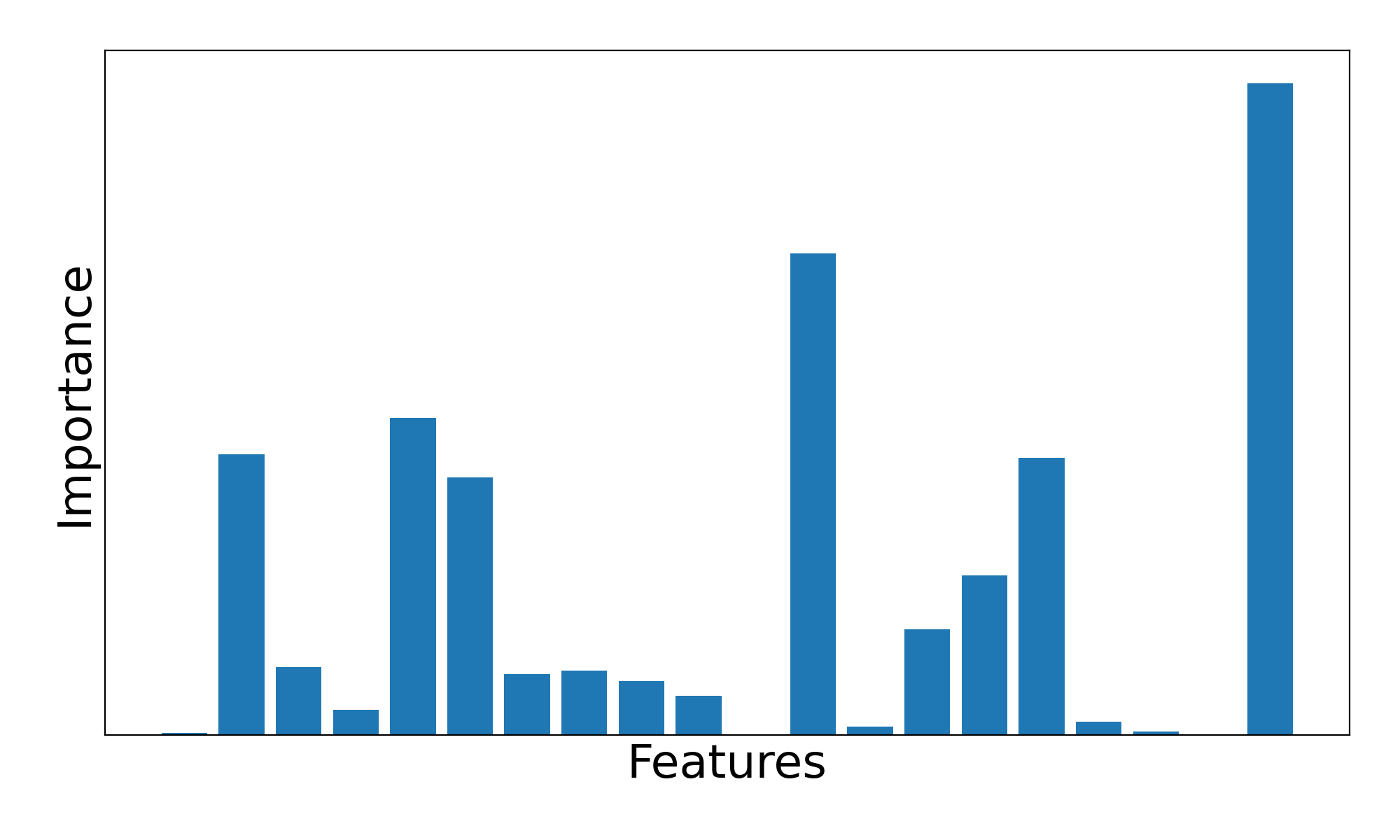}
         \caption{User identification}
         \label{fig:Identification_Features}
     \end{subfigure}
    \caption{Feature importance in two tasks}
    % \vspace{-7mm}
    \label{fig:Feature_Importance}
\end{figure}

Despite the advantages and potentials of these technologies, a critical challenge remains in selecting the most appropriate features for different applications while preserving user privacy. Often, features that are highly discriminative for one application, such as activity recognition, may inadvertently facilitate user identification~\cite{Nguyen2022}, thereby compromising privacy~\cite{Vanhamme2019IMUSensitiveData}. This duality presents a complex multi-objective optimization problem, where improving performance in one aspect may deteriorate another~\cite{Poli2021MOEAPrivacy}.
We observe that one feature subset may benefit one task while penalizing others.
For example, Figure~\ref{fig:Feature_Importance} shows the feature importance in two tasks: breathing activity recognition and user identification. For visualization purpose, we only select 20 features (see Section~\ref{sec:experiments} for the description of all features). The mean decreases in impurity as feature importance is calculated by a random forest classifier~\cite{Breiman2001}.
We observe that features which are important to one task may not benefit the other.
However, there are several features that are useful for both.
Hence, it is desirable to employ a feature selection procedure to enhance the performance of one task and reduce the accuracy of the other. This requires a multi-objective optimization algorithm to fulfil two or more criteria.
Traditional feature selection methods such as Recursive Feature Elimination (RFE)~\cite{Guyon2002} often focus on a single objective, such as maximizing classification accuracy. While effective in certain scenarios, these techniques are not designed to handle multiple conflicting objectives, and thus may not be suitable for applications that require a nuanced balance between different criteria, such as activity recognition and user privacy.
Other techniques such as Principal Component Analysis (PCA)~\cite{Jolliffe2016} focus on reducing the number of dimensions while preserving the maximum amount of information, regardless of the applications.
These existing methods generally tackle the optimization requirements in isolation, leading to sub-optimal solutions that fail to address the multifaceted requirements of real-world applications~\cite{Zhou2011}.

In contrast, our work employs a bio-inspired multi-objective genetic algorithm~\cite{Zhou2011} for feature selection. The evolutionary nature of this algorithm allows for a more adaptive and flexible feature selection process, capable of optimizing multiple objectives simultaneously. This is particularly advantageous in RF sensing applications, where a single sensor setup can capture a multitude of features with varying degrees of relevance and sensitivity to different tasks~\cite{Nguyen2022}. By utilizing a population of potential solutions and iteratively refining them, the multi-objective genetic algorithm provides a more comprehensive exploration of the feature space compared to traditional methods, thus enabling a more nuanced trade-off between conflicting objectives.
Our methodology is validated on a dataset comprising 50 subjects engaged in four distinct breathing activities, demonstrating a substantial divergence in accuracy between activity recognition and user identification. By focusing on this multi-objective framework, our work contributes to the ongoing discourse on ethical and effective human sensing, paving the way for more responsible and versatile applications of RF sensing technologies.
Our contribution includes:
\begin{itemize}
    \item We formulate the feature selection process as a multi-objective optimization problem. The chosen features based on vital signs can be used to implement breathing activity recognition with high accuracy while limiting the performance of sensitive tasks such as user identification.
    \item We evaluate our proposed approach on a novel dataset of 50 subjects, performing four distinct breathing patterns in two positions. One of the pattern is breath-holding, which simulates apnea or temporary cessation of breathing.
    \item We explore parameters for the genetic algorithm: the population size and the number of generations. In addition, we propose using efficient classification models on reduced inputs to calculate the fitness function faster.
\end{itemize}

Our proposed approach can be customized to satisfy the users' requirements in health monitoring.
For example, one system can offer high accuracy in breathing activity recognition but does not reveal the user identity.
Another application can focus on detecting the presence of one specific user but does not expose the one's activities.
In addition, selecting optimal feature subsets can lower the computation cost of training and employing the machine learning models.

%%%%%%%%%%%%%%%%%%%%%%%%%%%%%%%%%%%%%%%%%%%%%
%                                           %
%               RELATED WORK                %
%                                           %
%%%%%%%%%%%%%%%%%%%%%%%%%%%%%%%%%%%%%%%%%%%%%
\section{Related Work}
\label{sec:related}

Feature selection techniques play a pivotal role in improving the performance and interpretability of machine learning models, and they can be broadly classified into three categories: Filter Methods, Wrapper Methods, and Embedded Methods \cite{pudjihartono2022review}. Filter methods are computationally efficient and easy to implement but often ignore the interactions between features, thus potentially leading to suboptimal solutions. These methods rely on statistical tests to evaluate the relevance of each feature independently. Examples include the Chi-Squared Test, which measures the dependence between individual features and the target variable, and metrics such as Information Gain and Mutual Information that quantify the informativeness of a feature relative to the target. Correlation Coefficients evaluate the linear relationship between each feature and the target, while Variance Thresholding aims to exclude low-variance features under the premise that they have minimal predictive power \cite{Remeseiro2019FSMedicalFilterMethods}. In contrast, Wrapper Methods, although computationally more intensive, take into account feature interactions and often lead to better performance \cite{Xue2016WrappedFSReview}. These methods treat feature selection as a search problem, evaluating the merit of different feature subsets based on the performance of a predictive model. Recursive Feature Elimination (RFE) starts with all features and iteratively removes the least impactful ones based on model performance until a predefined number of features is attained \cite{Guyon2002}. Sequential Forward Selection (SFS) and Sequential Backward Selection (SBS) employ a greedy strategy to add or remove features based on a specific criterion \cite{ferri1994comparative}. Significantly, Genetic Algorithms stand out in this category by utilizing evolutionary strategies to optimize feature subsets, making them particularly relevant for multi-objective optimization tasks in feature selection \cite{pudjihartono2022review}. This is crucial when balancing conflicting objectives, such as maximizing classification performance while minimizing model complexity. Finally, Embedded Methods offer the benefit of inherently including feature selection during the model training process, albeit being algorithm-specific \cite{Guo2019EmbeddedFS}. Lasso Regression employs L1 regularization to force some coefficients to zero, effectively performing feature selection but potentially over-penalizing and eliminating relevant features. Ridge Regression uses L2 regularization and while it does not completely nullify features, it offers a way to gauge their importance. Elastic Net amalgamates the features of both L1 and L2 regularization. Decision Trees and Random Forests intrinsically provide feature importances, and neural networks can be tailored with regularization terms in their loss functions to serve as embedded feature selectors \cite{Guo2019EmbeddedFS}.

Understanding the advantages and drawbacks of these feature selection methods aids practitioners in making informed choices for their specific modeling tasks. Among these, the role of Genetic Algorithms~\cite{Zhou2011} is noteworthy for its ability to address multi-objective optimization in feature selection~\cite{Barbiero2019}~\cite{Poli2021MOEAPrivacy}~\cite{Njoku2023}, a key consideration when multiple conflicting objectives are at play. This makes Genetic Algorithms a significant topic of discussion in the related work, especially when balancing complex trade-offs in feature selection criteria is essential.
Previous work has leverage multi-objective optimization techniques to perform feature selection~\cite{Barbiero2019}~\cite{Poli2021MOEAPrivacy}~\cite{Njoku2023}.
They selected features for one classification problem with multiple objectives such as model complexity and performance metrics.
Barbiero~\textit{et al.}~\cite{Barbiero2019} considered three objectives: minimizing the feature subset size, minimizing the test error, and maximizing the analysis of variance value of the lowest-performing feature in the set.
Njoku~\textit{et al.}~\cite{Njoku2023} utilized two performance metrics (accuracy and area under the curve) of a classifier to formulate feature selection as a multi-objective optimization problem.
Poli~\textit{et al.}~\cite{Poli2021MOEAPrivacy} utilized two objectives: maximizing the accuracy of activity recognition and pushing gender recognition probability close to random guessing.
While their approach could limit gender recognition, the accuracy of activity recognition was reduced from 89.59\% to 84.14\%.
A similar setting to our case is multitask learning~\cite{Caruana1997} which aims to improve the accuracy of multiple classifiers by jointly training them.
Our approach selects features that benefit one classification model while penalizing another.

In order to reduce the exposure of identity details, Abbasi~\textit{et al.}~\cite{Abbasi2022} blurred photos at various degrees, which corresponded to the users' privacy requirement. They formulated an function to represent the relationship between privacy and accuracy of face recognition frameworks.
Through a user survey on important visual privacy features, Wang~\textit{et al.}~\cite{Wang2023} introduced a method for modeling the trade-off of privacy preservation and activity recognition on low-resolution images.
Following this line of research, Nguyen~\textit{et al.}~\cite{Nguyen2023} not only applied blurring but also other transformation such as adding noise, eyemask, and facemask to the images to deteriorate video quality.
The authors showed that these transformation techniques affect the rPPG signals.
After that, they applied restoration methods to enhance the vital signs measurement while still decreasing the perceived quality metrics.
These approaches modified the raw data before extracting features, which might influence the performance of all classification models trained on the altered inputs.
On the contrary, our approach strives to find a feature subset to improve the performance of one specific task.

%%%%%%%%%%%%%%%%%%%%%%%%%%%%%%%%%%%%%%%%%%%%%
%                                           %
%               METHODOLOGY                 %
%                                           %
%%%%%%%%%%%%%%%%%%%%%%%%%%%%%%%%%%%%%%%%%%%%%

\section{Multi-objective Feature Selection}
\label{sec:methodology}

\begin{figure}
\centering
    \includegraphics[width=0.85\columnwidth]{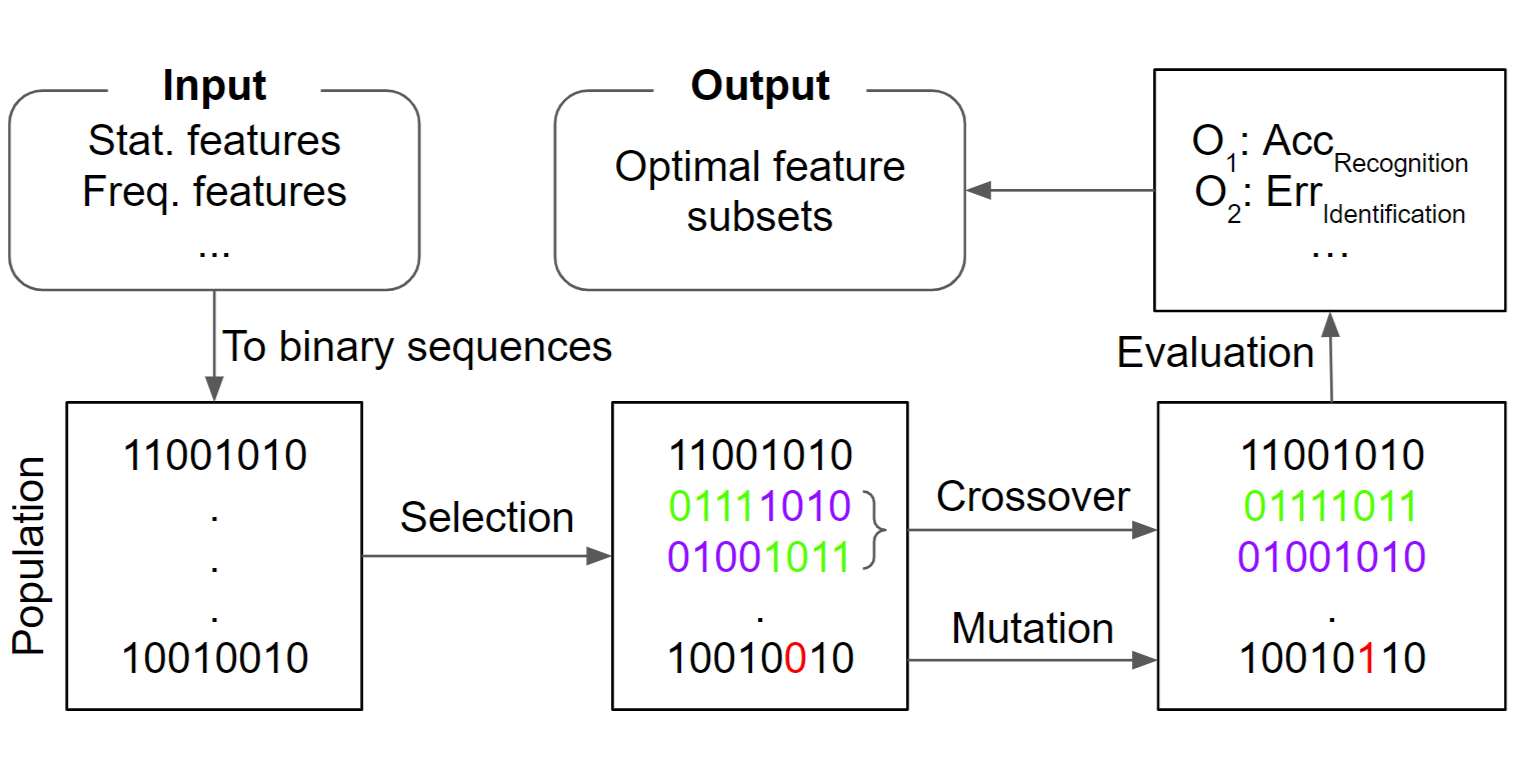}
         \caption{Multi-objective bio-inspired feature selection}
         \label{fig:Feature_Selection}
\end{figure}

In this section, we describe the feature engineering process and the classification models used in the proposed approach.
From the signals exported by the radar, we extract features describing vital signs, physiological attributes, morphological characteristics, entropy-based indicators, and fractal dimensions.
Then, our multi-objective optimization method selects the optimal feature subsets to train and evaluate two classification models: one for recognizing the breathing patterns (see Figure~\ref{fig:activities_distribution_radar}) and one for identifying the users.
Figure~\ref{fig:Feature_Selection} illustrates the genetic algorithm to select the best feature subset according to multiple objectives.

\subsection{Feature Extraction}
\begin{figure}
 \begin{center}
   \includegraphics*[width=0.49\textwidth]{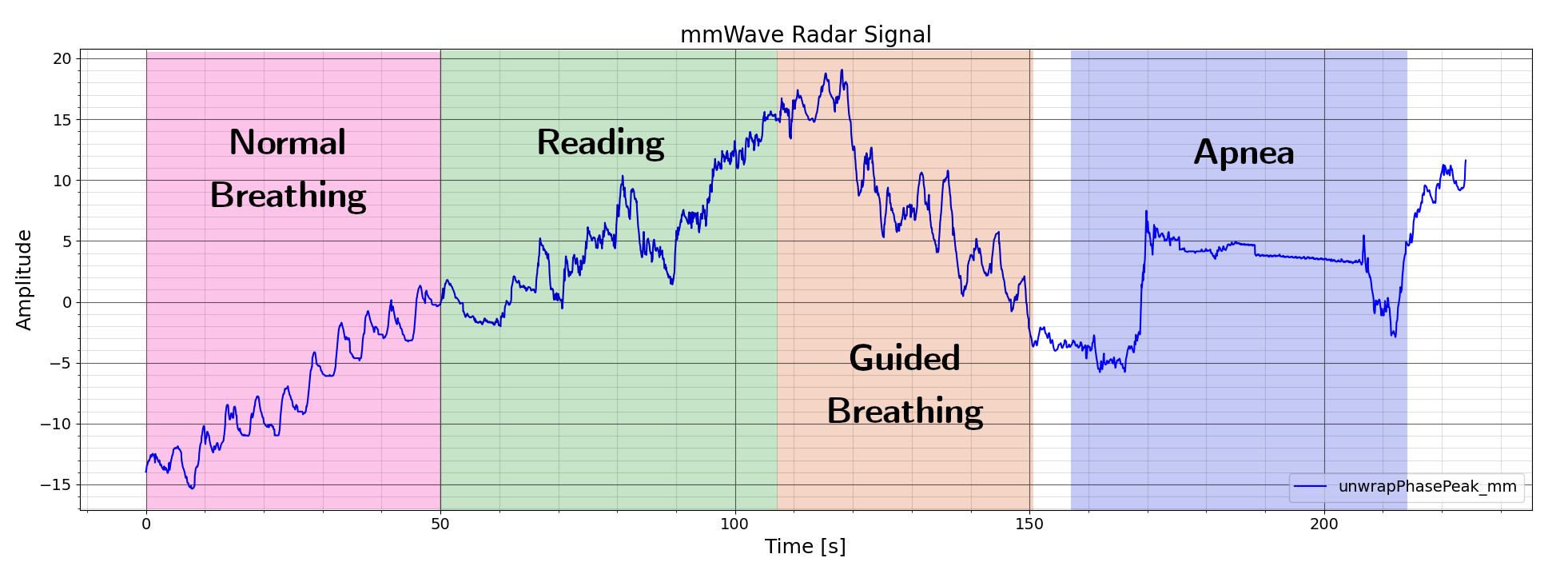}
 \end{center}
 \vspace{-3mm}
 \caption{Four different breathing activities captured with an mmWave radar.}
 \label{fig:activities_distribution_radar}
 % \vspace{-3mm}
\end{figure}

The mmWave radar performs the preliminary preprocessing phase to extract the vital signs, including respiratory and cardiac signals.
It also provides the chest displacement data.
For example, the respiratory waveforms are displayed in Figure~\ref{fig:activities_distribution_radar}, which are annotated with distinct breathing patterns.
We segment these signals into 10-second windows through a systematic windowing procedure, a crucial element in our data pipeline. The windows are shifted by a one-second interval, leading to an overlap rate of approximately 90\%. This overlapping rate is intentionally chosen to mitigate information loss during segmentation.

\begin{figure}
 \begin{center}
   \includegraphics*[width=0.49\textwidth]{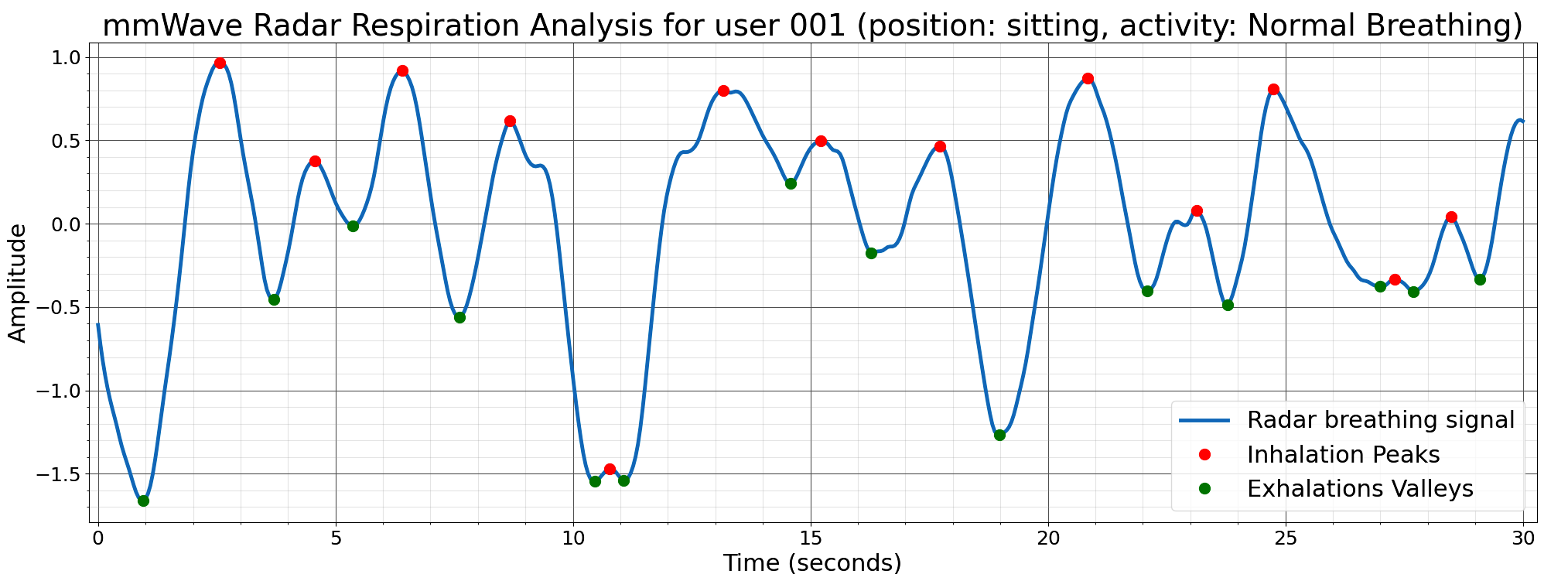}
 \end{center}
 % \vspace{-3mm}
 \caption{Detected peaks in the breathing signals: we extract Respiratory Rate Variability using the detected peaks.}
 \label{fig:brv_radar_analysis}
 % \vspace{-3mm}
\end{figure}

Subsequent to windowing, the methodology focuses on an extensive feature extraction phase, moving beyond conventional statistical attributes that are usually confined to temporal and frequency domains, following a similar approach proposed by Alvarez~\textit{et al.}~\cite{Alvarez2023DepressionRPPGs}. The extracted features span a broad array of parameters, including vital signs, physiological attributes, morphological characteristics, entropy-based indicators, and fractal dimensions. The feature set comprises 189 attributes extracted from the three signals associated with each user. The statistical parameters include basic statistics such as mean, minimum, maximum, and standard deviation, along with dynamic range, signal-to-noise ratio, median, and quartiles.
These features have been widely used in radio-frequency-sensing activity recognition~\cite{Sigg2013}.
They are further supplemented by time-domain and frequency-domain attributes like slope, total spectral power, and energy. Fractal analysis includes Katz, Higuchi, and Petrosian fractal dimensions, among others. Entropy features include permutation and spectral entropy, along with approximate and sample entropy. In the domain of heart and respiration features, our methodology embraces the computation of up to 70 physiological attributes, predominantly focusing on Heart Rate Variability (HRV)-related indicators. These HRV-related attributes span time-domain, frequency-domain, and non-linear-domain categories, meticulously detailed in previous references. Time-domain HRV features center around inter-beat intervals (IBI), time intervals between successive heartbeats, and normal beat intervals (NN intervals). The respiratory indicators are focused on the Respiratory Rate Variability (RRV), including parameters related to the peaks detected in the inhalations and exhalations (see Figure~\ref{fig:brv_radar_analysis}). Custom features, specifically designed to address breathing patterns and the unique differences between facial and thoracic biosignals, are also introduced, leveraging both time and frequency domains when applicable.

For the feature extraction purpose, these Python libraries are utilized: NumPy~\cite{Harris2020}, Antropy\cite{AntropyPythonPackage}, NeuroKit2 \cite{Makowski2021neurokit}, and HeartPy \cite{HeartPyVANGENT2019}. Our feature extraction process culminates in a multidimensional and discriminative feature set, pivotal for the precise identification and interpretation of physiological conditions within the framework of our feature selection method.

\subsection{Multi-objective Optimization for Feature Selection}
\label{sec:featureselection}

Feature selection plays a critical role in enhancing machine learning models' performance and interpretability. It involves the critical task of choosing a subset of relevant features from a given dataset while discarding irrelevant or redundant ones. This process not only simplifies the model but also mitigates overfitting risks, leading to improved generalization capabilities. However, feature selection is a computational demanding process because the search space increases exponentially with the number of features~\cite{Guyon2003}.
The number of feature subsets $s$ containing $d$ features for a dataset with total $N$ features is:
$\begin{aligned}
\sum _{d=1}^{N}{N \atopwithdelims ()d} = 2^{N} - 1
\end{aligned}$.
In our case, it is even more challenging since we aim to optimize multiple objectives.

With the same input features, we train two classification models: one to recognize the users' breathing activities and the other to identify the users.
We derive three objectives to find the optimal features for the first model and those features decreases the performance of the second model.
\begin{itemize}
    \item The first objective $O_1$ is to maximize the accuracy $a_R$ of the breathing activity recognition model: 
    \begin{align}
    \label{eq:O1}
        O_1 = a_R = \frac{1}{K} \sum_{i=1}^{K} \delta(y_i, \hat{y}_i),
    \end{align}
    where $K$ is the total number of recognized samples, $y_i$ is the ground-truth label for sample $i$, $\hat{y}_i$ is the predicted label for sample $i$, and $\delta(y_i, \hat{y}_i) = 1$ if $y_i = \hat{y}_i$ and $0$ otherwise.
    \item The second objective $O_2$ is to maximize the error rate of the user identification task:
    \begin{align}
    \label{eq:O2}
        O_2 = 1 - a_I,
    \end{align}
    where $a_I$ is the accuracy of the user identification model. 
    \item The third objective $O_3$ is to maximize the performance disparity between two model:
    \begin{align}
    \label{eq:O3}
        O_3 = a_R - a_I
    \end{align}
\end{itemize}
The multi-objective optimization problem is formulated as:
\begin{equation}
    \hat{s} = \mathop {\mathrm{argmax}} \limits _{s  \subset {\mathcal {F}}} (O_1, O_2, O_3),
\end{equation}
where $\mathcal {F}$ is the set of features.

We apply Non-dominated sorting genetic algorithm (NSGA-II)~\cite{Deb2002} to search for the optimal feature subset.
NSGA-II generates solutions using such biologically-inspired operators as crossover and mutation.
Each solution is a binary sequence corresponding to the features.
In these sequences, a value of 1 indicates that the feature is selected. 
Then, it selects the next generation with non-dominated sorting and crowding distance comparison.
In our expriments, we use the implementation provided in the library Inspyred from Tonda~\cite{Tonda2020}.

NSGA-II~\cite{Deb2002} (see Algorithm~\ref{alg:nsgs-ii}) combines non-dominated sorting, crowding distance assignment, and genetic operators (crossover and mutation) to efficiently search for a diverse set of solutions in multi-objective optimization problems.
A solution is  non-dominated, or Pareto-optimal, if none of its objective functions can be improved without degrading some of the other objective values.
Formally, let $x_1$ and $x_2$ be two solutions (i.e., binary sequences representing the selected features).
$x_1$ dominates $x_2$ if and only if $O_i(x_1) \geq O_i(x_2)$ for all $i \in \{1, 2, \dots , K\}$ and $O_j(x_1) > O_j(x_2)$ for at least one $j \in \{1, 2, \dots , K\}$. % (denoted by $x_1 \prec x_2$)
A Pareto front contains the set of Pareto-optimal solutions.
NSGA-II sorts the population into multiple non-dominated fronts.
The first front contains all Pareto optimal solutions, which are non-dominated by all other solutions in the solution space.
Then, all solutions in the first front are removed and the second front is calculated.
The procedure is repeated for the $n^{th}$ front through removing all the points from the first to the $(n-1)^{th}$ fronts and finding the Pareto optimal solutions from the remaining solutions.
The parents are selected from these fronts with the decreasing probability.
NSGA-II conducts these following steps to produce the archive $A$ containing non-dominated solutions:
\begin{itemize}
    \item \textbf{Step 1: Initialization} This step initializes a population of individual solutions, often randomly or through some heuristic method, and calculates their fitness functions (see Algorithm~\ref{alg:fitness}).
    \item \textbf{Step 2: Selection} The second step selects individuals from the current population to form the parent set for the next generation,  based on its dominance rank and crowding distance (used for diversity preservation). NSGA-II typically uses a binary tournament selection mechanism. Two individuals are randomly chosen, and the one with higher dominance rank and greater crowding distance is selected as a parent. This procedure is repeated until the parent population size is filled.
    \item \textbf{Step 3: Crossover} The third step applies crossover operators to pairs of parent solutions to create offspring.
    \item \textbf{Step 4: Mutation} This step applied mutation operators to the offspring population to introduce diversity. The mutation is typically applied with a certain probability for each parameter of each offspring (i.e., flipping between selecting and deselecting a feature).
    \item \textbf{Step 5: Combine Parent and Offspring Populations} This step combines the parent and offspring populations to create a merged population for the next generation.
    \item \textbf{Step 6: Fast Non-dominated Sorting} The sixth step performs a fast non-dominated sorting of the merged population to divide individuals into multiple fronts or Pareto fronts. It assigns a rank to each individual based on its front membership, with individuals in the first front having rank 1.
    \item \textbf{Step 7: Crowding Distance Assignment} This step calculates the crowding distance for each individual within each front. The crowding distance measures the density of individuals around a specific solution in the solution space. It encourages diversity by prioritizing solutions that are not too close to each other.
    \item \textbf{Step 8: Select Next Generation} This step selects individuals for the next generation by first taking all individuals from the first front (rank 1) and then adding individuals from subsequent fronts until the population size is reached. When adding individuals from a front, it prioritizes those with higher crowding distances. The Pareto-optimal solutions are appended to the archive $A$. 
    \item \textbf{Step 9: Termination} The steps from 2 to 8 are repeated for a predefined number of generations or until a termination criterion is met (e.g., a predefined number of generations).
\end{itemize}

The mathematical formulations mainly involve defining the dominance relationship and crowding distance calculation. Denote an individual as $u$ and $v$ and their objectives as $O^u$ and $O^v$.
The crowding distance calculation for sorting within fronts contains these steps for each objective function $O_i$:
\begin{itemize}
    \item Sort the individuals based on their values of $O_i$.
    \item Assign a crowding distance to each individual as the difference in $O_i$ values of the adjacent individuals.
    \item The boundary individuals get infinite crowding distances to ensure they are selected.
\end{itemize}

The effectiveness of NSGA-II~\cite{Deb2002} depends on specific problems, fitness functions, and parameter settings. It balances the exploration and exploitation of the objective space to find a diverse set of Pareto-optimal solutions in the archive $A$.

\begin{algorithm}
\caption{NSGA-II: Non-Dominated Sorting Genetic Algorithm II}
\label{alg:nsgs-ii}
\begin{algorithmic}[1]

\Procedure{NSGA-II}{}
\State Initialize population $P$ with random individuals
\State Initialize archive $A$ as an empty set
\State Set generation count $t$ to 1

\While{$t < \text{max\_generations}$}
    \State Evaluate the fitness of each individual in $P$
    \State Perform non-dominated sorting to create fronts $F_1, F_2, \ldots, F_k$
    \State Calculate the crowding distance of individuals in each front
    \State Create a new population $Q$ by selecting individuals from $P$ based on fronts and crowding distance
    \State Apply genetic operators (crossover and mutation) to individuals in $Q$
    \State Perform non-dominated sorting on $Q \cup P$ to create new fronts $F'_1, F'_2, \ldots, F'_k$
    \State Select the next population $P$ from $Q \cup P$ by filling it with individuals from $F'_1, F'_2, \ldots,$ until the population size is reached
    \State Update the archive $A$ with non-dominated solutions from $P$
    \State Increment $t$ by 1
\EndWhile

\State \textbf{return} Pareto front in $A$

\EndProcedure

\end{algorithmic}
\end{algorithm}

\begin{algorithm}
\caption{Fitness function}
\label{alg:fitness}
\begin{algorithmic}[1]

\Procedure{Fitness function}{}

    \State Train the breathing activity recognition model $M_1$
    \State Train the user identification model $M_2$
    \State Calculate the accuracy of $M_1$: $O_1$ (see Equation~\ref{eq:O1})
    \State Calculate the error rate of $M_2$: $O_2$ (see Equation~\ref{eq:O2})
    \State Calculate the performance disparity between $M_1$ and $M_2$: $O_3$ (see Equation~\ref{eq:O3})
    \State \textbf{return} $O_1, O_2, O_3$
\EndProcedure

\end{algorithmic}
\end{algorithm}

%%% End of ChatGPT <-- Done modification but can still polish

%% TODO: LE commented this before integrating RFE comparison <-- DONE
For comparison purposes as well as demonstrating the validity of our approach, we also explore a popular feature selection method known as Recursive Feature Elimination (RFE)~\cite{Guyon2002}. RFE is a popular choice in the data science community for its simplicity and effectiveness. It offers ease of configuration and implementation, making it accessible to both novice and experienced users. Its primary objective is to identify the most pertinent features within the dataset, those that significantly influence the prediction of the target variable. By iteratively removing less impactful features, RFE efficiently narrows down the feature set, ultimately resulting in enhanced model performance and increased interpretability. RFE possesses notable strengths, including its ease of use and effectiveness in identifying relevant features. It simplifies the feature selection process while aiding in reducing overfitting risks. However, it is not without limitations. RFE can be computationally intensive, particularly with large datasets. Furthermore, its performance may be influenced by the choice of the underlying model. While RFE is a valuable tool, it may not always identify the globally optimal feature subset due to its stepwise elimination approach.

\subsection{Classification Models}

In the current study, we focus on one ensemble learning algorithms that are grounded in decision trees, namely Random Forest (RF)~\cite{Breiman2001} for the classification tasks. These tree-based ensemble models provide multiple advantages, including the ability to assess feature importance. In Random Forest, feature importance is commonly determined through evaluating the mean decrease in impurity when a specific feature is perturbed, while holding other features constant. This yields a relative importance score for each feature, thus allowing the algorithm to prioritize highly impactful variables. This is particularly useful for high-dimensional datasets.
Extremely Randomized Tree~\cite{Geurts2006} is another ensemble learning that can be utilized. Unlike Random Forest, which employs a deterministic approach to selecting feature split points based on impurity reduction, Extremely Randomized Trees add an extra layer of randomness by choosing both the feature and the threshold for splitting randomly. This additional randomness mitigates the model's potential for overfitting at the expense of slightly increasing its bias. Both algorithms, Random Forest and Extremely Randomized Trees, build each base estimator using a random subset of features, enhancing the diversity among individual trees in the ensemble. Overall, the differentiation between the two methods lies primarily in their approach to node splitting: Random Forest optimizes feature splits based on impurity metrics, while Extremely Randomized Trees introduce additional randomness in the threshold selection process.
Both of them leverage a majority voting mechanism among the trees to produce the final predictions.

Another classification and regression that relies on majority voting is the k-nearest neighbours algorithm (k-NN)~\cite{Cover1967}.
From the class labels of the $k$ nearest neighbours in the feature space, k-NN makes a prediction for a given sample.
In order to do that, the model computes the similarity between data points in the feature space, using different metrics such as Euclidean distance and cosine similarity.
For classification tasks, k-NN utilizes a majority vote to determine the class label of the new sample: the label that occurs most frequently among the $k$ neighbours is chosen as the predicted class.
For classification tasks based on RF signals, k-NN is a suitable method that balances accuracy and efficiency~\cite{Nguyen2022}.

The Python~\textit{Scikit-learn} library~\cite{scikit-learn} was utilized to implement all of the classifiers in our experiments.
Prior to the model training phase, we handle any anomalies in the extracted features, such as missing or out-of-range values. These anomalies are addressed either by replacing them with the preceding (or succeeding) valid value in the column or by imputing them with the column mean.
Then, we perform standardization to transform data of each column so that it has zero mean and unit variance.
These steps ensure that the features have a common scale before utilizing them in further experiments and analysis.

%%%%%%%%%%%%%%%%%%%%%%%%%%%%%%%%%%%%%%%%%%%%%
%                                           %
%           EXPERIMENTAL SETUP              %
%                                           %
%%%%%%%%%%%%%%%%%%%%%%%%%%%%%%%%%%%%%%%%%%%%%
\section{Experiments and Analysis}
First, this section introduces our novel dataset which contains chest displacement, respiratory signals, and cardiac signals of 50 subjects in two positions.
Second, it explains the evaluation protocol of our approach.
Next, we present our experiment results and analyze them to show the effectiveness of our feature selection method.

\label{sec:experiments}

% \subsection{Dataset}
\subsection{Benchmark Database}

We compiled a dataset consisting of 50 healthy subjects, maintaining an equal gender distribution (1:1 ratio), following the university procedure for scientific research participants and the EU General Data Protection Regulation.
The data acquisition process was clearly explained to all participants, who then signed the consent forms.
The age range of the participants were 24 - 65, with a mean age of 32.7 years old.
Their mean height was 169.5cm, their mean weight was 69.5kg, and their mean chest circumference was 0.93m.

\begin{figure}
    \centering
     \begin{subfigure}[b]{0.4\columnwidth}
         \centering
         \includegraphics[width=\columnwidth]{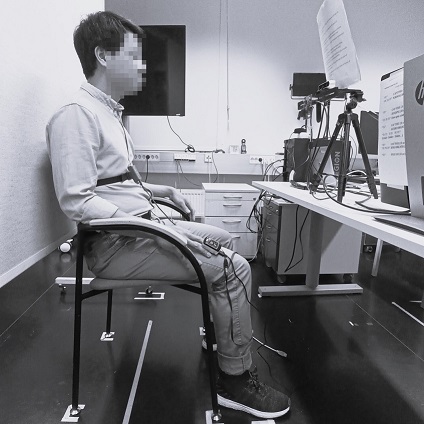}
         \caption{Sitting}
         \label{fig:Sitting}
     \end{subfigure}
     \hspace{5mm}
     \begin{subfigure}[b]{0.4\columnwidth}
         \centering
         \includegraphics[width=\columnwidth]{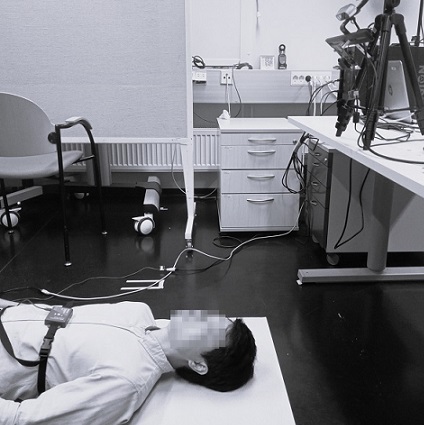}
         \caption{Lying}
         \label{fig:Lying}
     \end{subfigure}
    \caption{User positions in the data collection sessions}
    % \vspace{-7mm}
    \label{fig:Data_Collection}
\end{figure}

During the data collection sessions, each participant was seated and reclined, as illustrated in Figure~\ref{fig:Data_Collection}. In both positions, they performed four breathing patterns: normal breathing, reading a brief narrative, synchronized breathing following a predefined pattern (i.e., guided breathing), and breath-holding to simulate apnea (i.e., temporary cessation of breathing).
Each breathing activity lasted approximately 30 seconds as depicted in Figure \ref{fig:activities_distribution_radar}.
To capture chest displacement and derive respiration and heart waveforms, we used the Texas Instruments IWR1443 mmWave radar, sampling at a rate of 20Hz.
The distance between the chest area and the radar was 0.7m.
The vital signs produced by the radar outputs served as the inputs for our feature extraction and selection methods.

%%%%%%%%%%%%%%%%%%%%%%%%%%%%%%%%%%%%%%%%%%%%%
%                                           %
%               EXPERIMENTS                 %
%                                           %
%%%%%%%%%%%%%%%%%%%%%%%%%%%%%%%%%%%%%%%%%%%%%

\subsection{Parameters of the genetic algorithm}
\label{sec:GA_parameters}
We split 50 subjects into three groups for training (42 subjects), validation (4), and testing (4).
Note that identifying users in a four-member group is in accordance with previous work of Yang~\textit{et al.}~\cite{Yang2020}.
Moreover, increasing the group size decreases the overall performance~\cite{Nguyen2022}.
We selected the optimal features using their performance on the validation set.
For each task, we trained a random forest classifier~\cite{Breiman2001}.
The final evaluation was conducted on the testing set.
Note that we trained and tested the activity recognition model on different subjects.
On the other hand, the user identification model was trained on four testing subjects in the sitting position and tested on the same subjects in the lying position.

We varied the number of generations of the genetic algorithm~\cite{Deb2002} and observed that this parameter could affect the disparity between the two classifiers (see Table~\ref{tab:GA_Parameters}).
NSGA-II~\cite{Deb2002} was performed on the training set to generate candidate feature subsets.
The optimal subsets are used to train and evaluate the users on the testing data for two tasks: breathing pattern recognition and user identification.
The population size varies: 10, 20, and 50.
The number of generations varies: 30, 40, 50, 60, 80, and 100.
For each pair of parameters, we repeated the experiment 10 times and reported the average numbers.
We show the average accuracy in Table~\ref{tab:GA_Parameters}.
We observe that setting the population size to 50 and the number of generations to 50-80 balances the computational cost and the performance disparity between two tasks.

\begin{table}[]
\caption{Population size (Pop. size) and Number of generations (\#gen.) affecting the accuracy of two tasks}
\label{tab:GA_Parameters}
\centering
\begin{tabular}{|c|c|c|c|}
\hline
\textbf{Pop. size} & \textbf{\# gen.} & \textbf{Breathing Recognition} & \textbf{User Identification} \\ \hline
10 & 50     & 80.67\%       & 32\% \\ \hline
20 & 50     & 84.27\%       & 24.79\%  \\ \hline
50 & 30     & 82\%          & 34\% \\ \hline
50 & 40     & 83.97\%       & 28.72\% \\ \hline
50 & 50     & 85.18\%       & 30.62\% \\ \hline
50 & 60     & 86.19\%       & 30.36\%  \\ \hline
50 & 80     & 86.71\%       & 30.93\% \\ \hline
50 & 100    & 82.95\%       & 29.95\%  \\ \hline
\end{tabular}
\end{table}

\subsection{Comparison to the original feature set}

\begin{figure}
    \centering
     \begin{subfigure}[b]{0.48\columnwidth}
         \centering
         \includegraphics[width=\columnwidth]{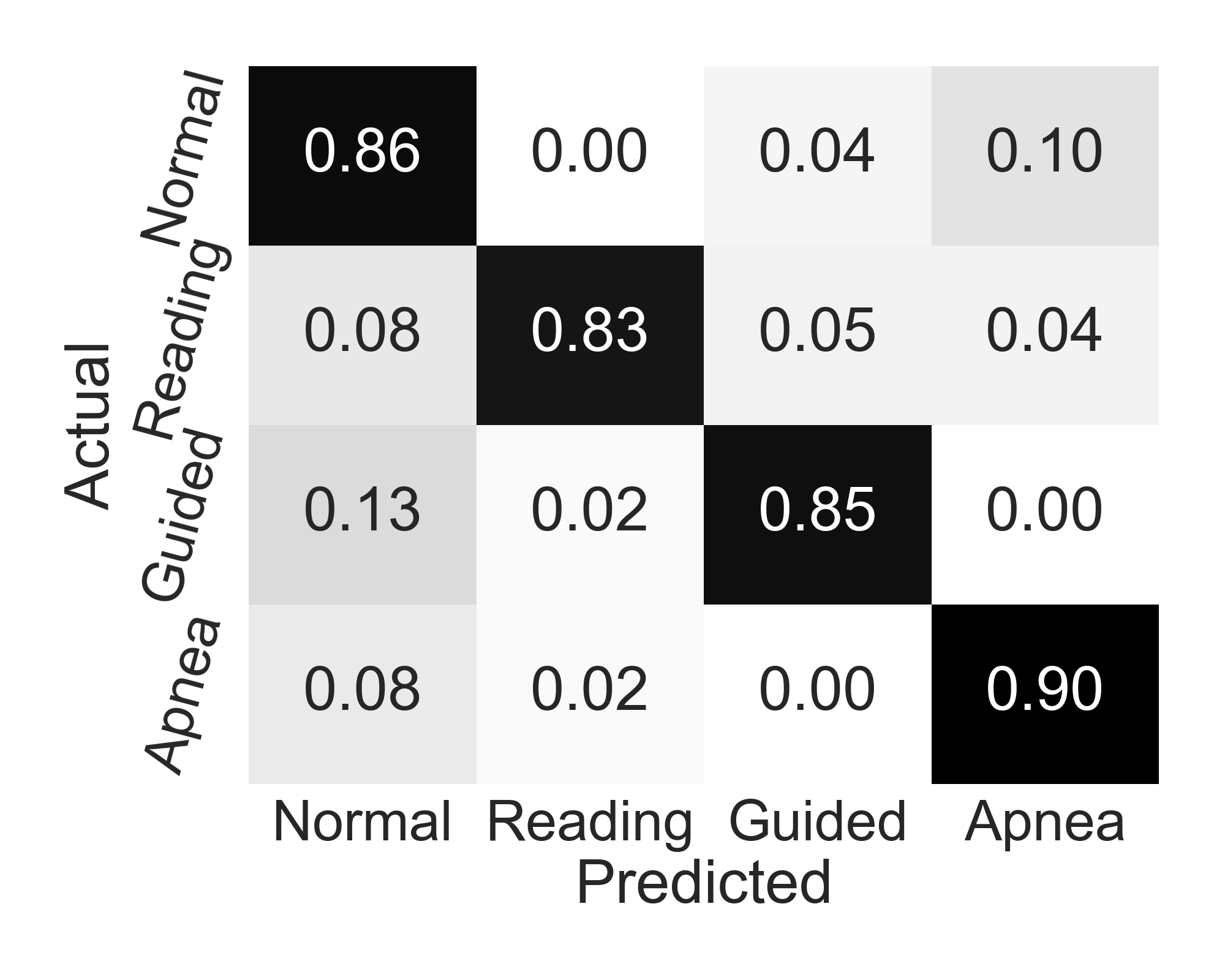}
         \caption{No feature selection (using all 189 features)}
         \label{fig:AR}
     \end{subfigure}
     \begin{subfigure}[b]{0.48\columnwidth}
         \centering
         \includegraphics[width=\columnwidth]{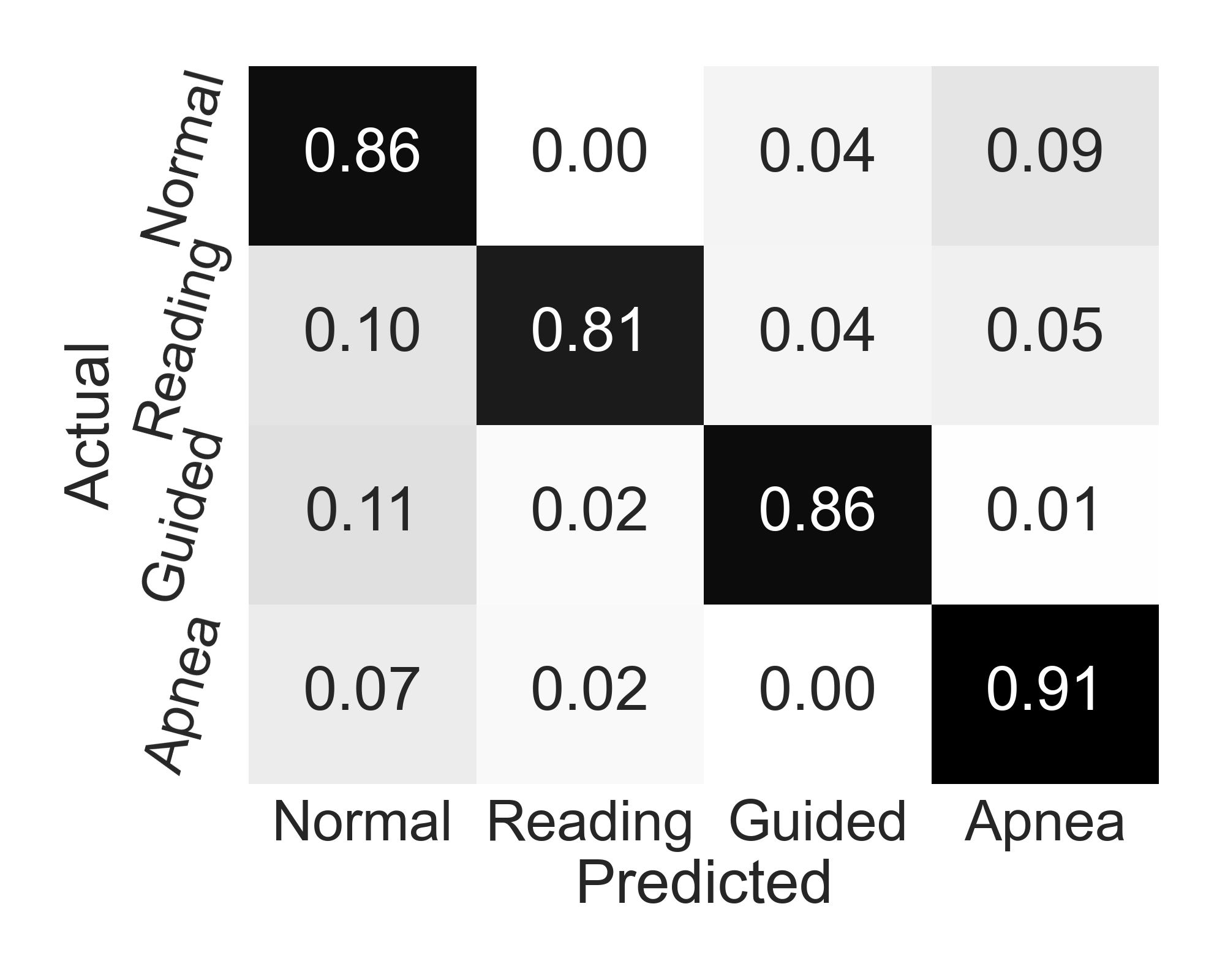}
         \caption{With the optimal features for breathing activity recognition}
         \label{fig:AR_FS}
     \end{subfigure}
    \caption{Breathing activity recognition}
    % \vspace{-7mm}
    \label{fig:AR_Performance}
\end{figure}

\begin{figure}
    \centering
     \begin{subfigure}[b]{0.48\columnwidth}
         \centering
         \includegraphics[width=\columnwidth]{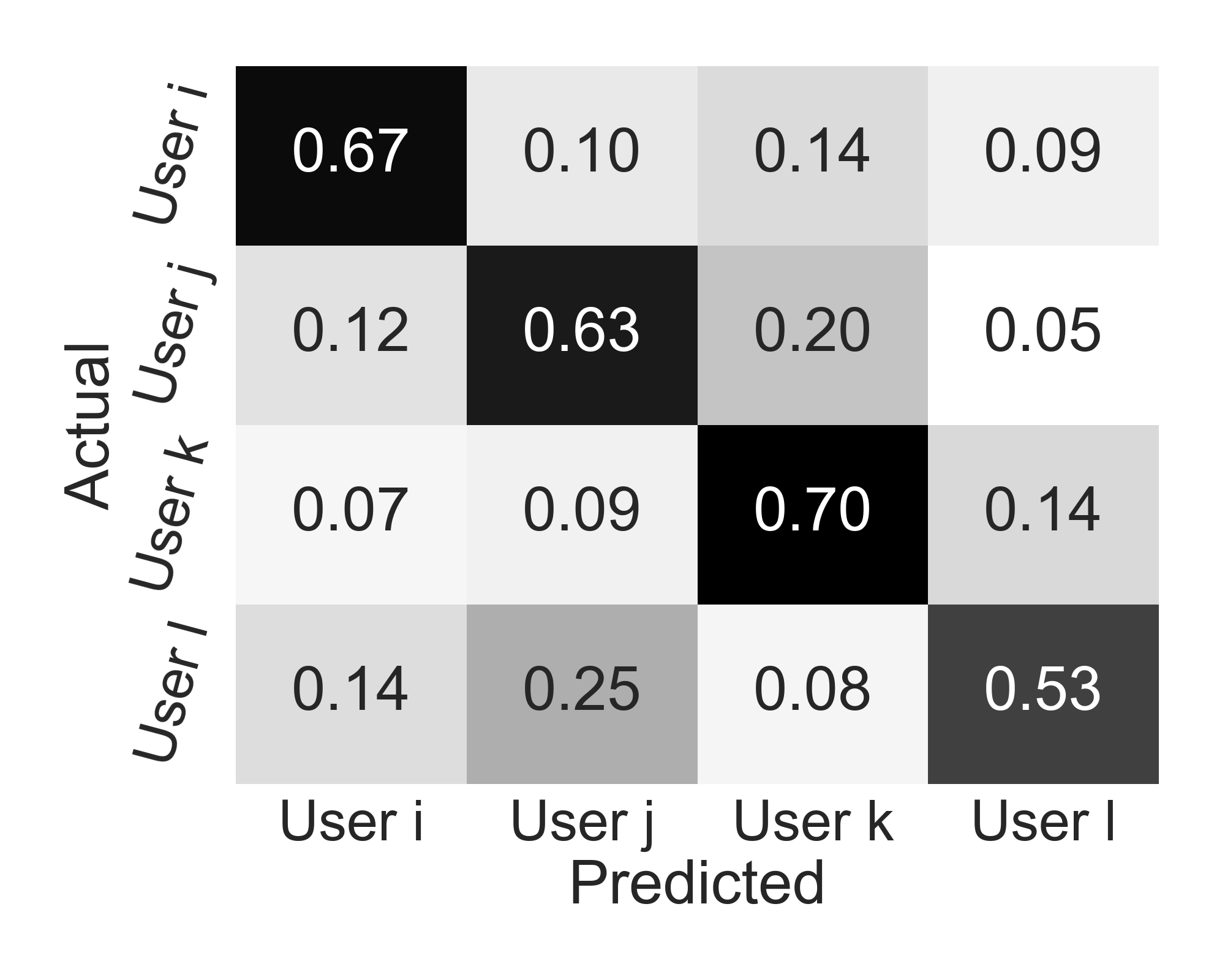}
         \caption{No feature selection (using all 189 features)}
         \label{fig:UI}
     \end{subfigure}
     \begin{subfigure}[b]{0.48\columnwidth}
         \centering
         \includegraphics[width=\columnwidth]{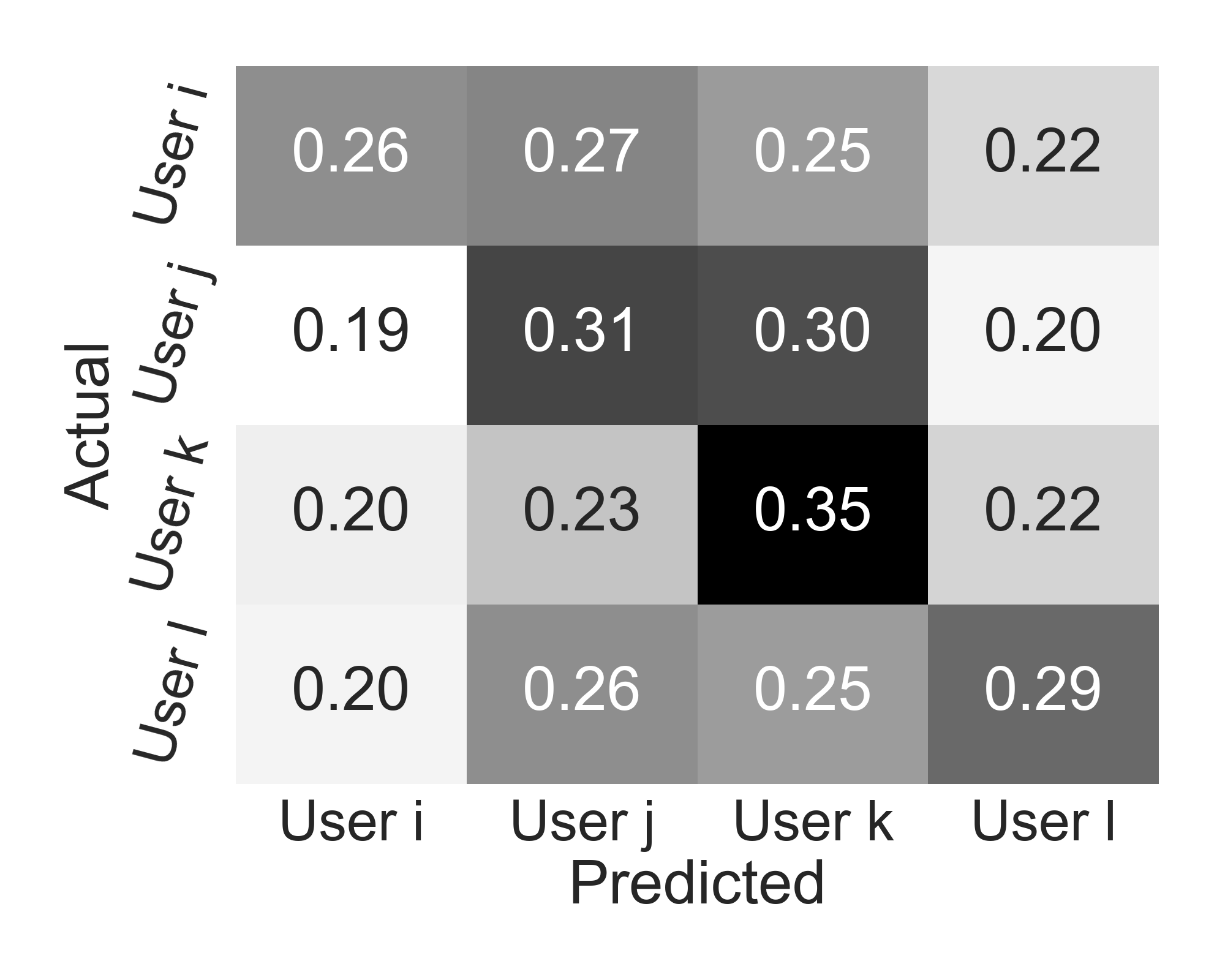}
         \caption{With the optimal features for breathing activity recognition}
         \label{fig:UI_FS}
     \end{subfigure}
    \caption{User identification}
    % \vspace{-7mm}
    \label{fig:UI_Performance}
\end{figure}

For comparison, we trained and evaluated the classification models on the original features set (189-dimension) in the leave-one-group-out (LOGO) setting: 46 subjects for training and 4 subjects for testing.
We repeated this experiment 20 times to obtain the average results.
In each turn, we randomly shuffled the set of subjects.
We used the optimal features for breathing activity recognition to reduce the dimensions of the dataset.
The subsets of optimal features were the best solutions found by the genetic algorithm with 80 generations and 50 candidates per generation (i.e., the population size).
Note that the best feature subsets was selected using the validation data (see Section~\ref{sec:GA_parameters} for more details).
Two RF classifiers were trained and assessed on the succinct data: one for breathing activity recognition and one for user identification.

In Figure~\ref{fig:AR_Performance}, after feature selection, the accuracy of recognizing breathing patterns was maintained.
On the original feature set, the mean accuracy was 85.98\% (standard deviation: 0.0521).
Using the optimal feature subset, the mean accuracy of breathing activity recognition was slightly improved to 86.07\% (standard deviation: 0.0523).
The breathing recognition model trained on the selected features confused between normal breathing, guided breathing, and apnea samples, as analogous to the model of all features.
Yet it improved the apnea recognition accuracy.
On the other hand, the performance of the user identification model was impaired by these optimal features (see Figure~\ref{fig:UI_Performance}).
Specifically, its accuracy reduced significantly from 63.23\% (standard deviation: 0.1407) to 30.18\% (standard deviation: 0.0697), which was closer to random guessing.
We conclude that our proposed approach benefits one task while hindering the other.

\subsection{Leave-one-subject-out evaluation}
\begin{figure}
    \centering
     \begin{subfigure}[b]{0.48\columnwidth}
         \centering
         \includegraphics[width=\columnwidth]{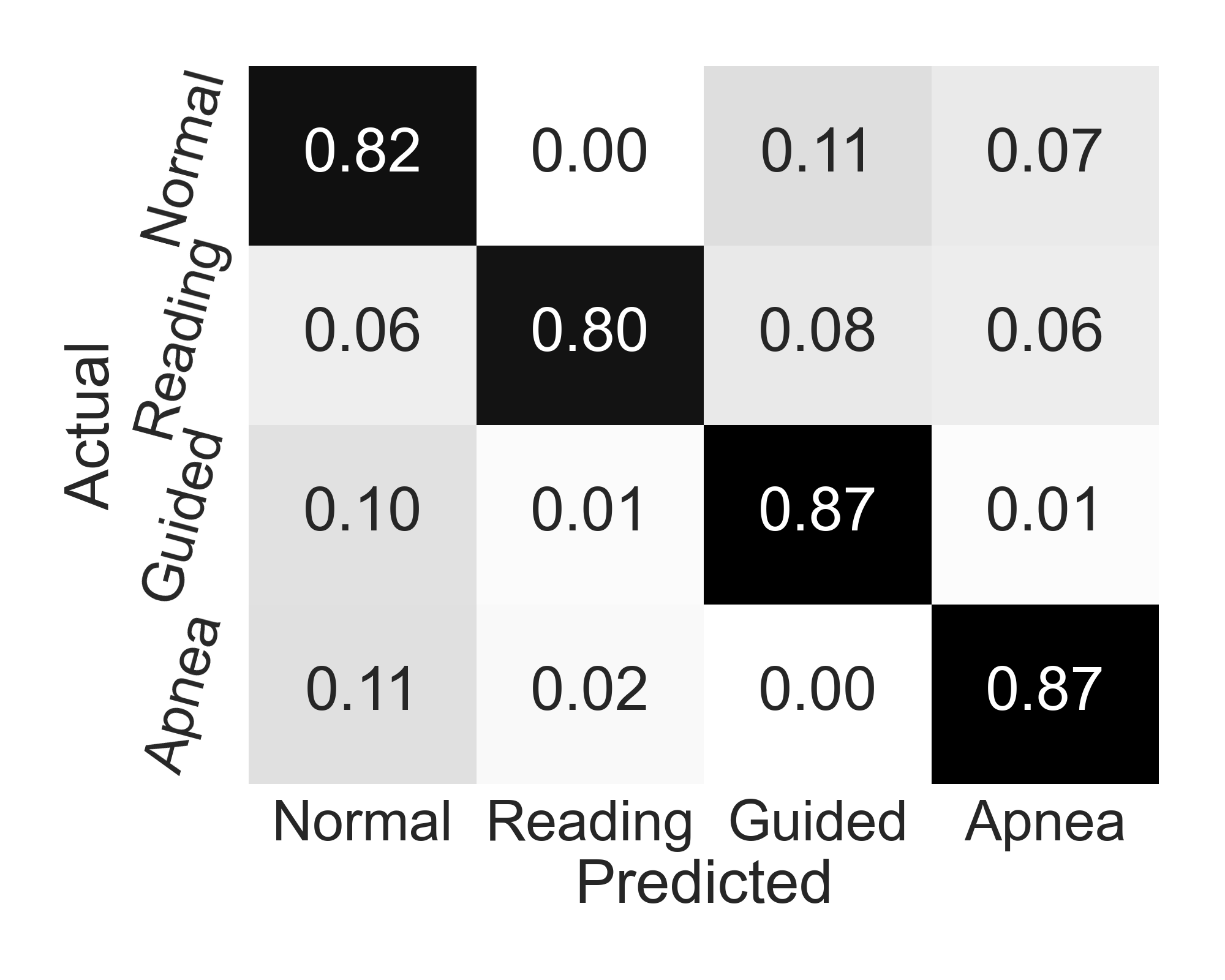}
         % acc 82%
         \caption{No feature selection}
         \label{fig:LOSO_AR_All_Features}
     \end{subfigure}
     \begin{subfigure}[b]{0.48\columnwidth}
         \centering
         \includegraphics[width=\columnwidth]{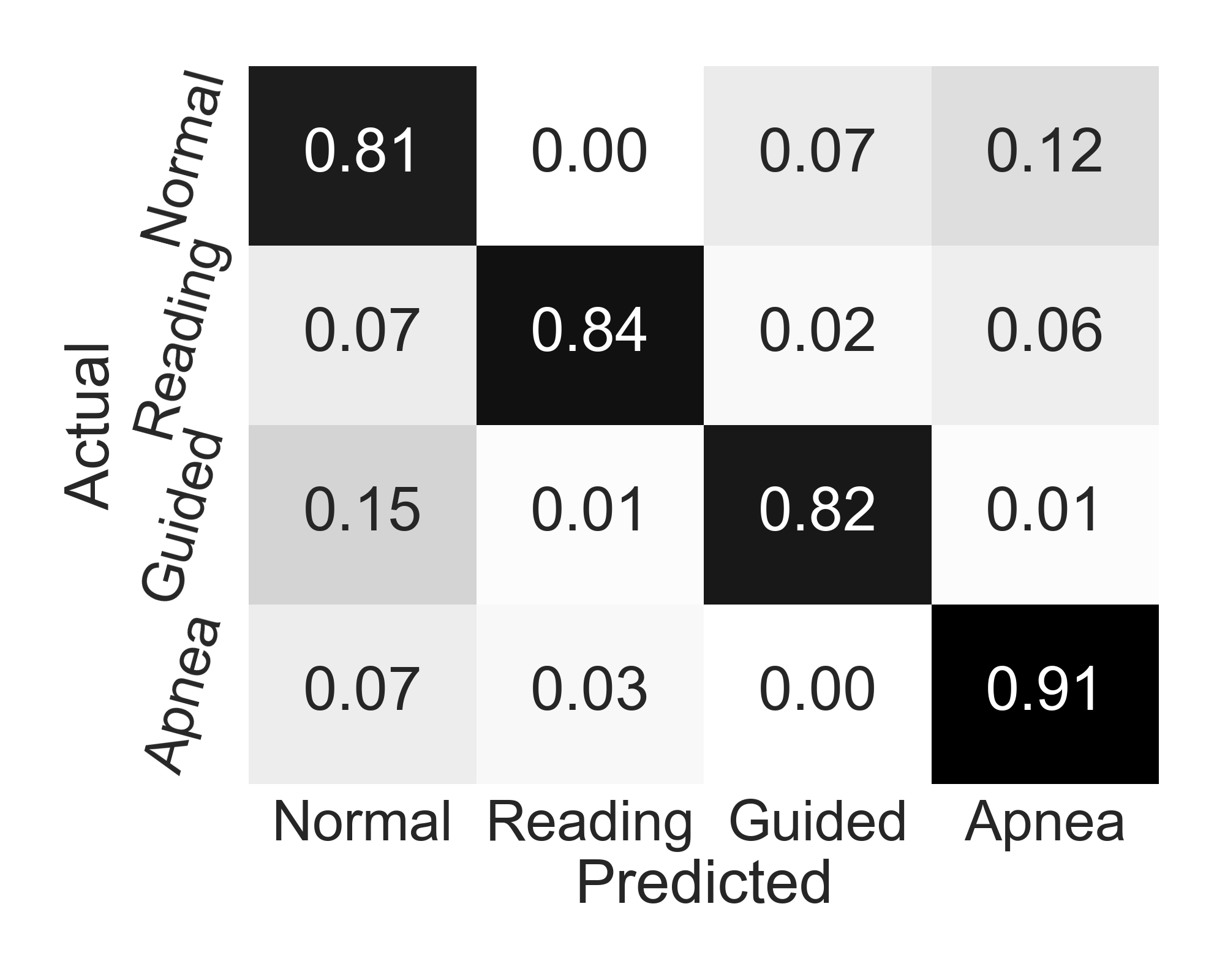}
         \caption{Optimal feature subset}
         \label{fig:LOSO_AR_Optimal_Features}
     \end{subfigure}
    \caption{Leave-one-subject-out breathing activity recognition}
    % \vspace{-7mm}
    \label{fig:LOSO_AR_Performance}
\end{figure}

In this experiment, we evaluated our method using a leave-one-subject-out (LOSO) cross-validation approach. For the activity recognition classification task, we trained 50 unique classifiers. In each case, one subject was intentionally excluded from the training set for testing. The accuracy metrics and confusion matrices were averaged to assess the model performance.
Figure~\ref{fig:LOSO_AR_Performance} show the confusion matrices of recognizing breathing activities on all features and selected features.
The classification model was a Random Forest with 100 estimators.
For the result of the feature selection method, we use the features selected by the genetic algorithm~\cite{Deb2002} with these settings: 50 as the population size and 80 as the number of generations.
The accuracy of the first model trained on the full dataset is 84.05\%.
That of the second model trained on the optimal feature subset for breathing activity recognition is 84.42\%.
Our proposed approach has maintained the performance while using fewer features.

\subsection{Simplifying the fitness function}

\begin{figure}
 \begin{center}
   \includegraphics*[width=0.48\textwidth]{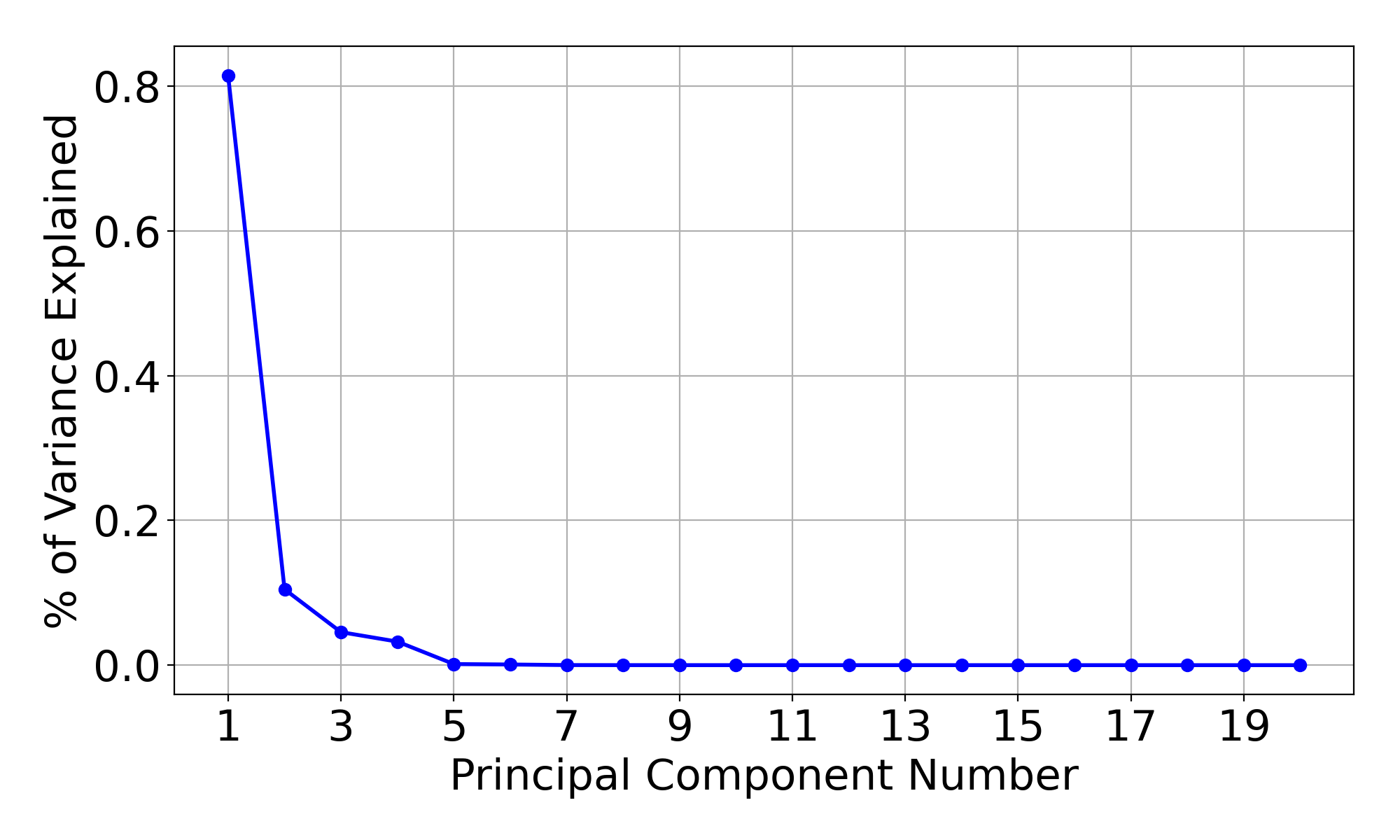}
 \end{center}
 \vspace{-3mm}
 \caption{PCA scree plot}
 \label{fig:pca_plot}
 % \vspace{-3mm}
\end{figure}

Computing the fitness function is costly since we need to train and evaluate two machine learning models.
Hence, we leverage two surrogate models to estimate the accuracy of breathing activity recognition and user identification (see Table~\ref{tab:PCA_Runtime}).
Specifically, we use PCA~\cite{Jolliffe2016} to reduce the number of dimensions of the input data.
We visualize the proportion of variance explained by the principal components to select the number of dimensions (see Figure~\ref{fig:pca_plot}).
The reduced number of dimensions is fixed to 5.
Then, we train and evaluate two k-NN models~\cite{Cover1967} on the reduced data ($k = 3$).
A k-NN classifier could balance accuracy and efficiency in RF-based classification problems, as empirically showed by Nguyen~\textit{et al.}~\cite{Nguyen2022}.
The runtime is measured on a desktop computer with Intel Core i5-12400F 2.50 GHz processor and 16 GB RAM.
Table~\ref{tab:PCA_Runtime} shows that we can significantly decrease the computing time of the fitness function while our proposed feature selection method still maintains a competitive performance.

\begin{table}[]
\caption{Runtime per generation of evaluating the full fitness function and the PCA-based fitness function (in brackets: accuracy of breathing activity recognition \& user identification)}
\label{tab:PCA_Runtime}
\begin{tabular}{|c|c|c|c|}
\hline
\textbf{Pop. size} & \textbf{\#gen.} & \textbf{Full fitness func.} & \textbf{PCA fitness func.} \\ \hline
 50 & 50 & 45s (85.18\% \& 30.62\%) & 15s (84.41\% \& 35.77\%) \\ \hline
 50 & 60 & 45s (86.19\% \& 30.36\%) & 15s (83.71\% \& 30\%) \\ \hline
 50 & 80 & 45s (86.71\% \& 30.93\%) & 17s (84.99\% \& 30.22\%) \\ \hline
 50 & 100 & 48s (82.95\% \& 29.95\%) & 15s (82.30\% \& 38.14\%) \\ \hline
\end{tabular}
\end{table}

\subsection{Single-objective feature selection}
We applied RFE~\cite{Guyon2002} to find the optimal feature subsets for breathing activity recognition.
Similar to the previous sections, we randomly selected 46 subjects for training and 4 for testing.
We used the optimal features of the breathing recognition classifier on user identification model.
We repeated the experiment 10 times to accumulate the average results.

\begin{figure}
    \centering
     \begin{subfigure}[b]{0.48\columnwidth}
         \centering
         \includegraphics[width=\columnwidth]{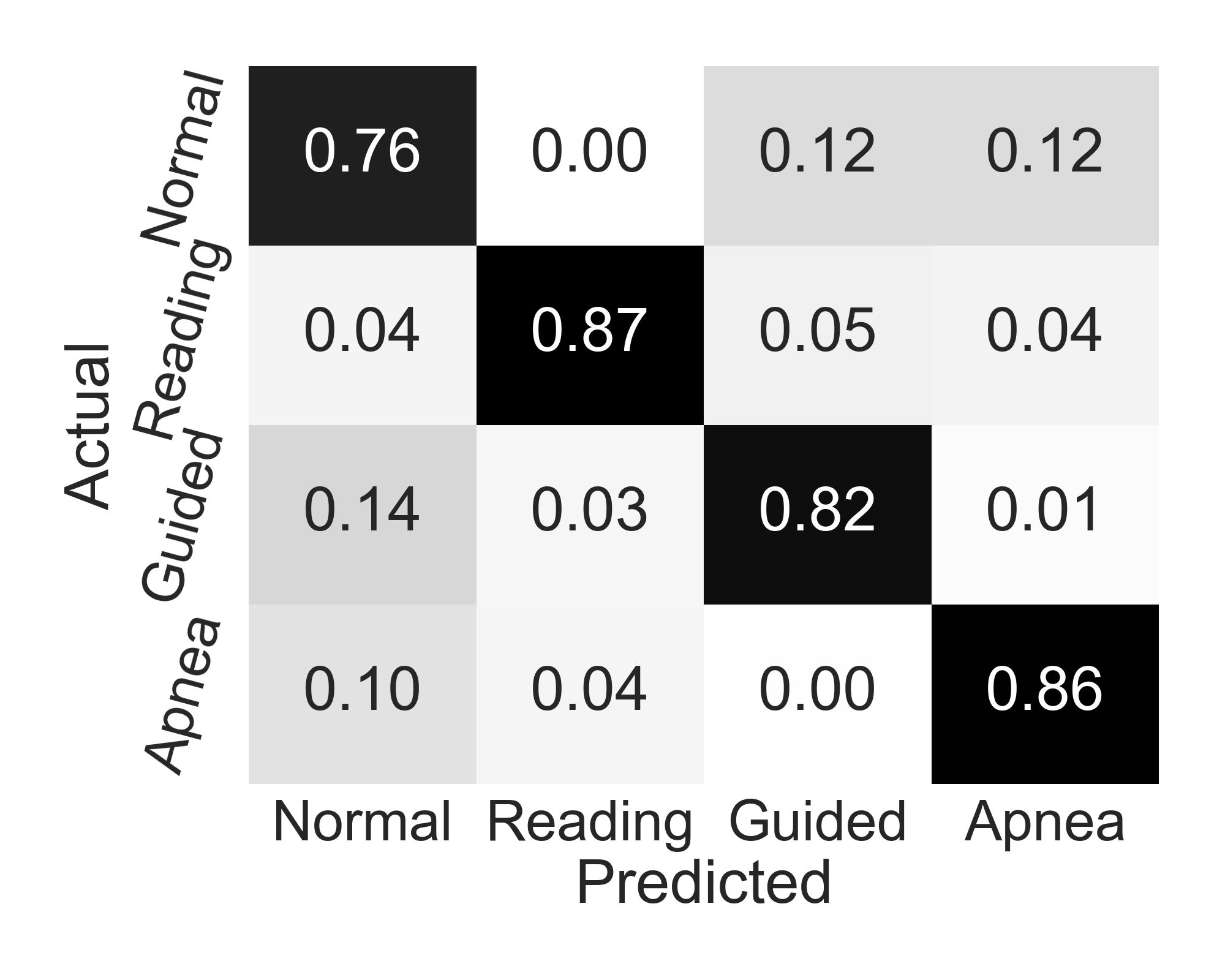}
         \caption{Breathing activity recognition}
         \label{fig:REF_LOGO_50_AR}
     \end{subfigure}
     \begin{subfigure}[b]{0.48\columnwidth}
         \centering
         \includegraphics[width=\columnwidth]{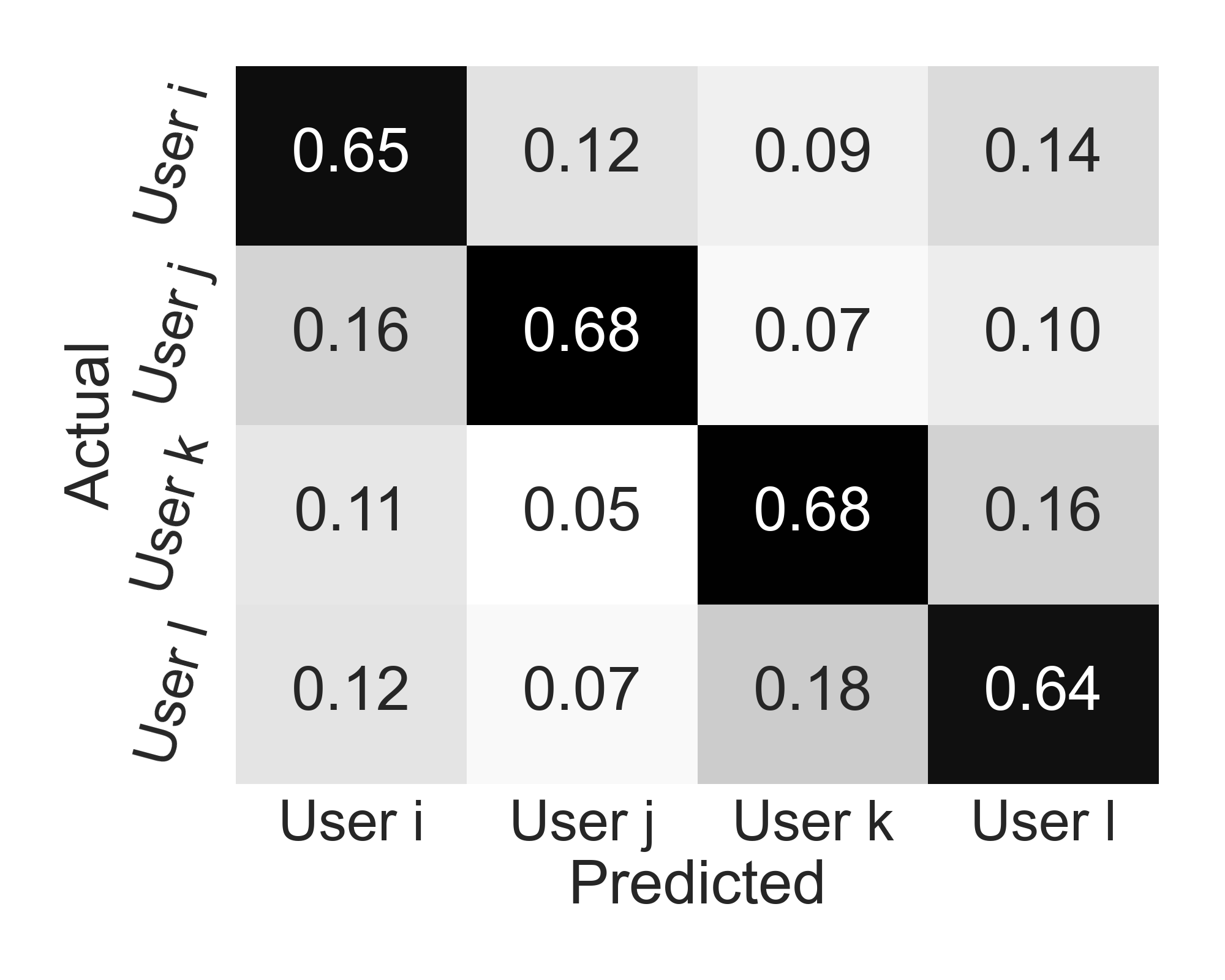}
         \caption{User identification}
         \label{fig:REF_LOGO_50_UI}
     \end{subfigure}
    \caption{Using RFE to obtain 50 features for breathing activity recognition and evaluating two models with these features}
    % \vspace{-7mm}
    \label{fig:RFE_LOGO_AR_50}
\end{figure}

\begin{figure}
    \centering
     \begin{subfigure}[b]{0.48\columnwidth}
         \centering
         \includegraphics[width=\columnwidth]{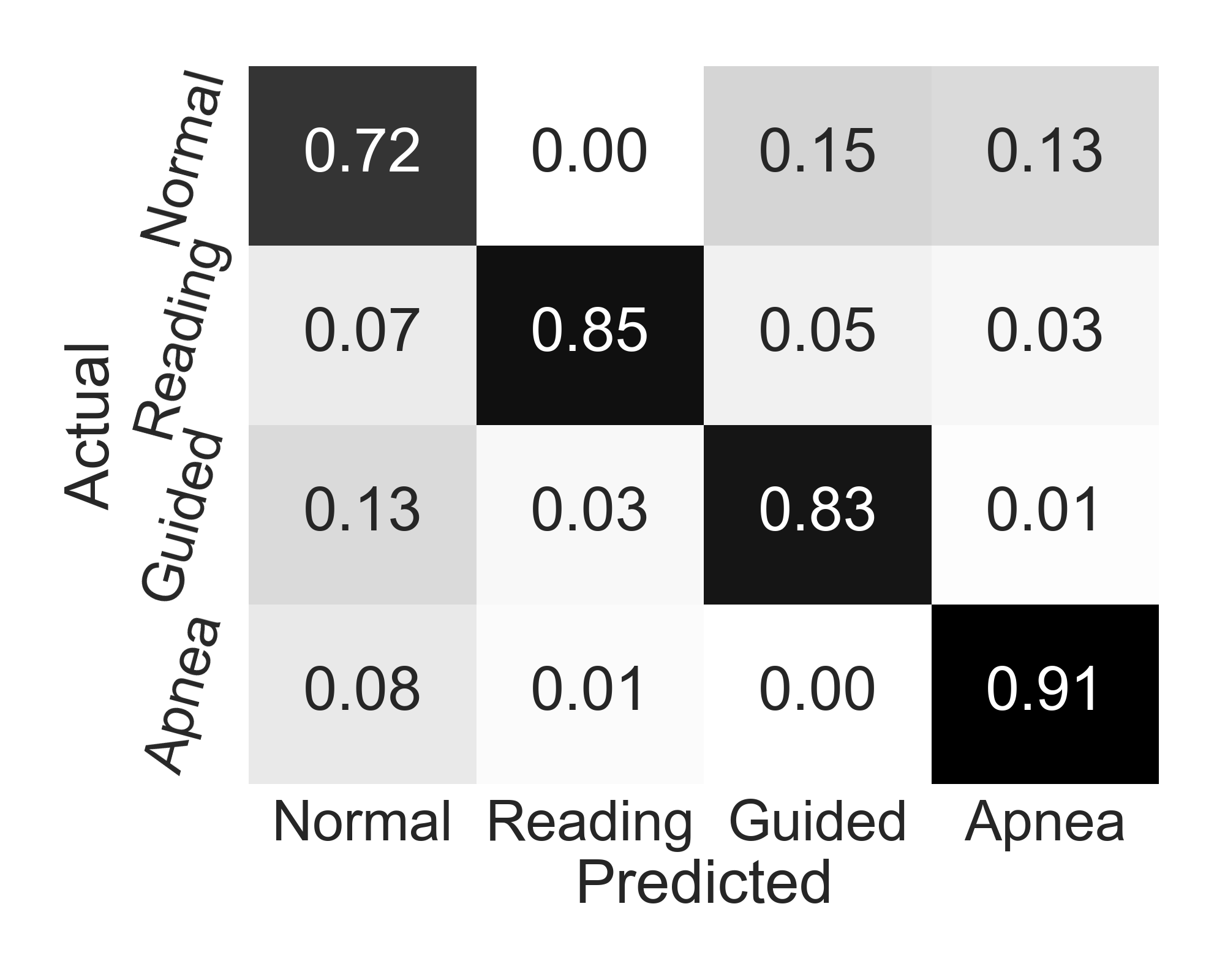}
         \caption{Breathing activity recognition}
         \label{fig:RFE_LOGO_100_AR}
     \end{subfigure}
     \begin{subfigure}[b]{0.48\columnwidth}
         \centering
         \includegraphics[width=\columnwidth]{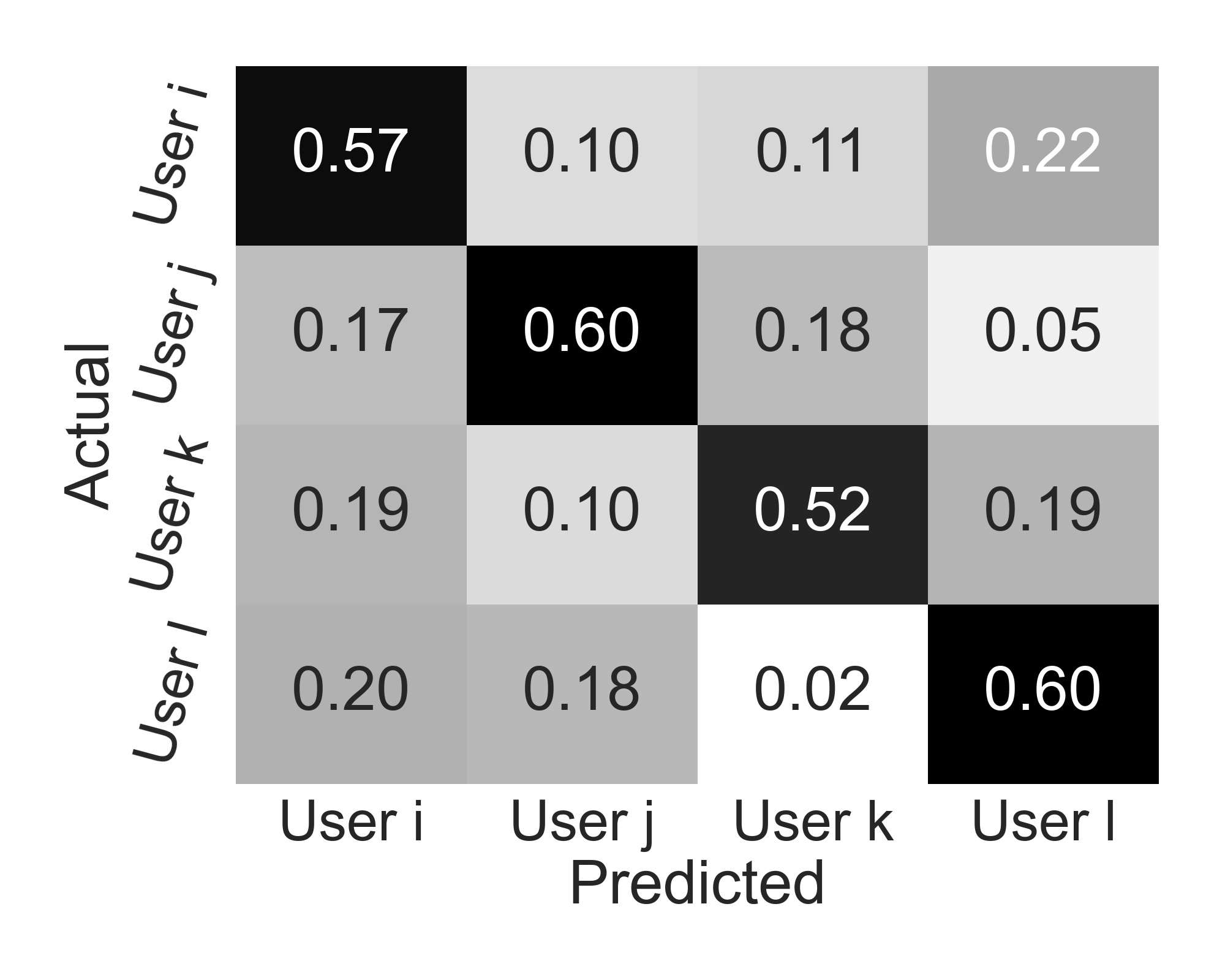}
         \caption{User identification}
         \label{fig:RFE_LOGO_100_UI}
     \end{subfigure}
    \caption{Using RFE to obtain 100 features for breathing activity recognition and evaluating two models with these features}
    % \vspace{-7mm}
    \label{fig:RFE_LOGO_AR_100}
\end{figure}

First, we fixed the number of selected features.
Figure~\ref{fig:RFE_LOGO_AR_50} and Figure~\ref{fig:RFE_LOGO_AR_100} show the confusion matrices when the number of features is 50 and 100, respectively.
With 50 features, the accuracy of breathing recognition was 82.54\% and that of user identification was 66.15\%.
With 100 features, our breathing recognition model achieved an accuracy of 82.51\% and the other model reached 57.32\%.
We can see that the accuracy of user identification is still high, comparing to our multi-objective feature selection approach.
This is due to the selected features, which are useful for the breathing recognition model, are beneficial to the user identification model.

\begin{figure}
    \centering
     \begin{subfigure}[b]{0.48\columnwidth}
         \centering
         \includegraphics[width=\columnwidth]{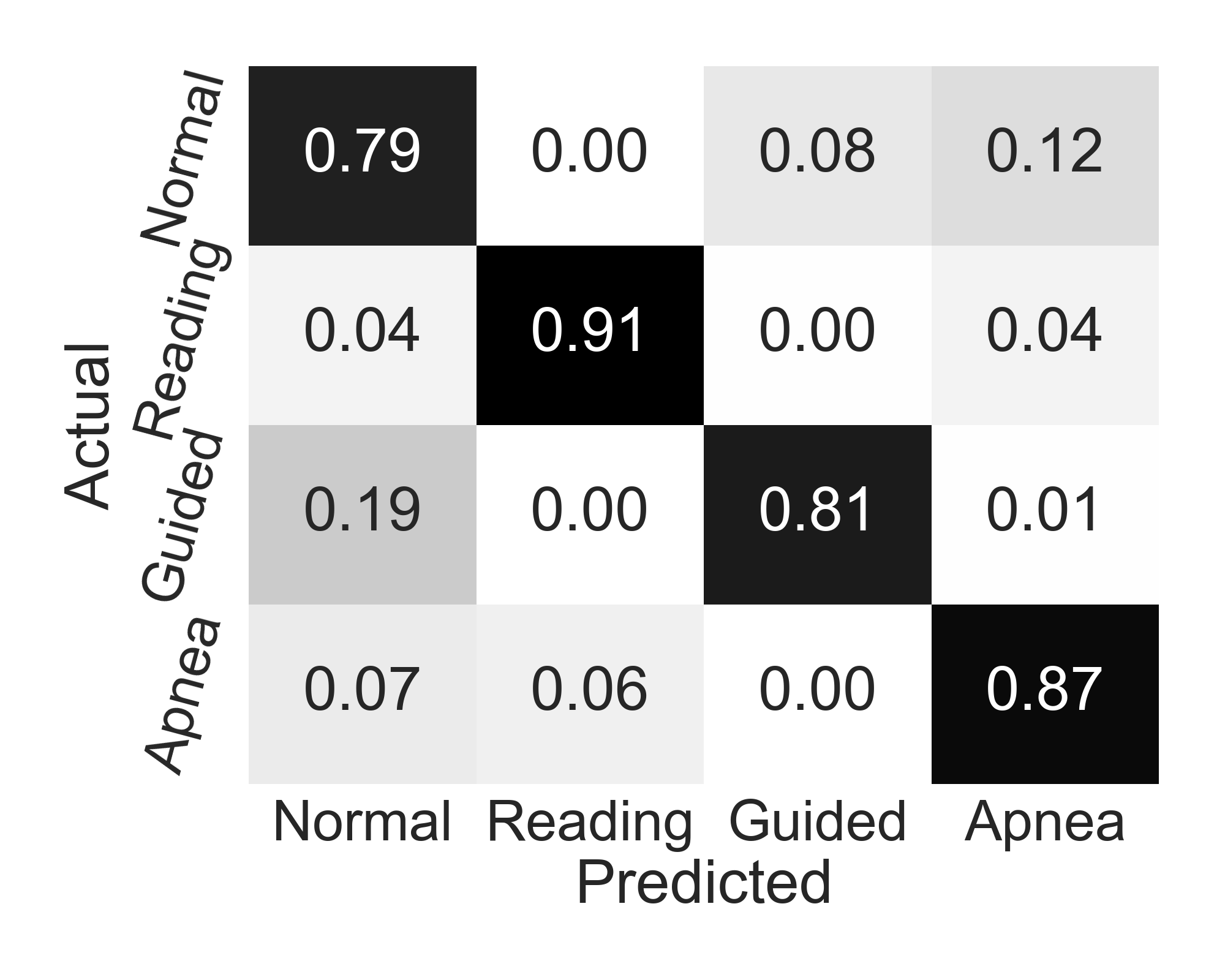}
         \caption{Breathing activity recognition}
         \label{fig:RFECV_LOGO_AR}
     \end{subfigure}
     \begin{subfigure}[b]{0.48\columnwidth}
         \centering
         \includegraphics[width=\columnwidth]{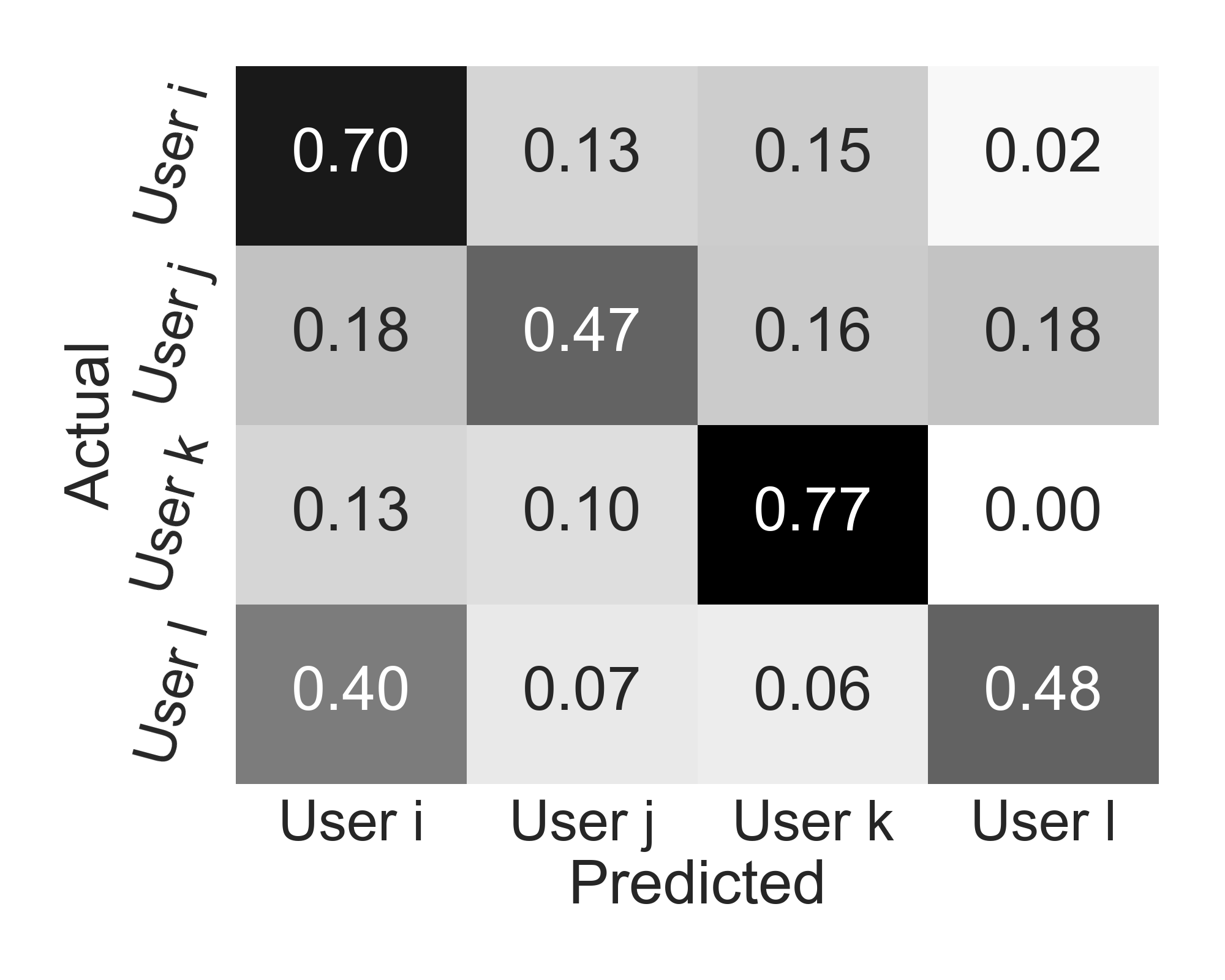}
         \caption{User identification}
         \label{fig:RFECV_LOGO_UI}
     \end{subfigure}
    \caption{Using RFE with cross validation to obtain the best features for breathing activity recognition and evaluating two models with these features}
    % \vspace{-7mm}
    \label{fig:RFECV_LOGO_AR_Performance}
\end{figure}

Second, we used a five-fold cross validation procedure to find the optimal number of features in each run.
The number of selected features ranged from 47 to 162.
We achieved a slightly higher accuracy for breathing activity recognition: 84.50\%.
However, the accuracy of user identification reached 60.49\%.
Figure~\ref{fig:RFECV_LOGO_AR_Performance} shows the confusion matrices of two models, using the optimal feature subsets for the first model.

From these two results, we conclude that selecting the best features for one model (i.e., single-objective feature selection) do not satisfy the requirement of improving one task and limiting others.
In addition, RFE removes presumably redundant features which may yield better class separation when being combined with others~\cite{Guyon2003}.
Hence, multi-objective methods are more suitable to find the optimal feature subsets in such scenarios as described in this article: we aim to develop an application to recognize breathing patterns but simultaneously we strive to hinder user identification.

\subsection{Best features for user identification}

In the previous sections, the emphasis has been on accurately identifying specific health-related activities, such as breathing patterns.
In this experiment, we reverse the objectives to find the best features for user identification.
Employing multi-objective optimization techniques, we aim to maximize user identification accuracy while simultaneously minimizing the system's capacity for activity recognition.
One potential use case is to detect the presence of a specific user without revealing one's breathing activities.

\begin{figure}
    \centering
     \begin{subfigure}[b]{0.48\columnwidth}
         \centering
         \includegraphics[width=\columnwidth]{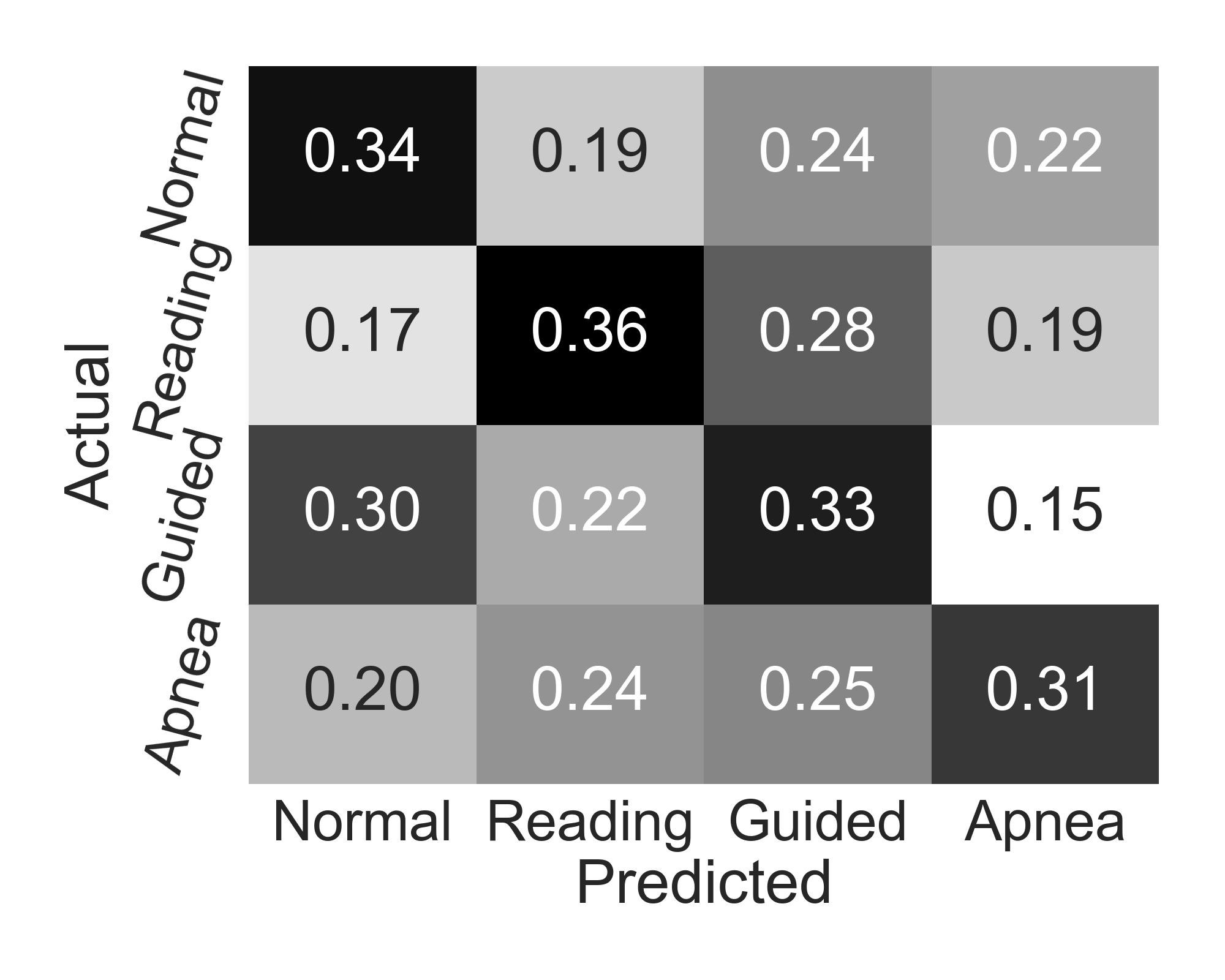}
         \caption{Breathing activity recognition}
         \label{fig:AR_Best_Features_UI}
     \end{subfigure}
     \begin{subfigure}[b]{0.48\columnwidth}
         \centering
         \includegraphics[width=\columnwidth]{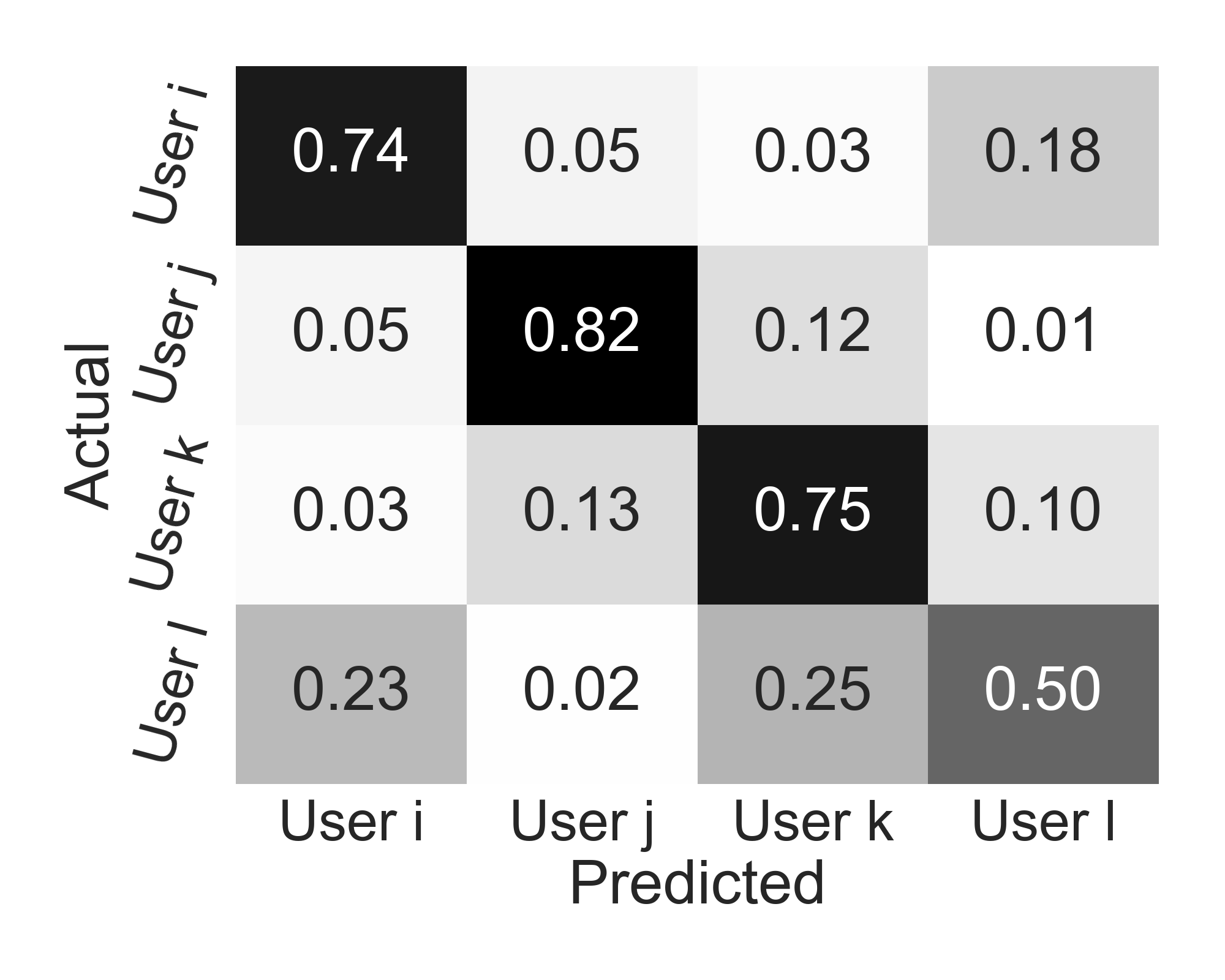}
         \caption{User identification}
         % acc 70%
         \label{fig:UI_Best_Features_UI}
     \end{subfigure}
    \caption{The optimal features selected for user identification improves this task while decreasing the performance of breathing activity recognition.}
    % \vspace{-7mm}
    \label{fig:UI_Best_Features}
\end{figure}

We evaluate the RF classifiers for activity recognition and user identification on these selected features.
Figure~\ref{fig:AR_Best_Features_UI} and Figure~\ref{fig:UI_Best_Features_UI} show the confusion matrices when performing the RF classifiers on activity recognition and user identification, respectively.
The optimal feature subset for identifying users improve the performance of this task (comparing Figure~\ref{fig:UI} using all features and Figure~\ref{fig:UI_Best_Features_UI} using only the best features for user identification) while decreasing the accuracy of recognizing the users' breathing patterns.

%%%%%%%%%%%%%%%%%%%%%%%%%%%%%%%%%%%%%%%%%%%%%
%                                           %
%               CONCLUSION                  %
%                                           %
%%%%%%%%%%%%%%%%%%%%%%%%%%%%%%%%%%%%%%%%%%%%%

\section{Conclusion}

\begin{figure}
 \begin{center}
   \includegraphics*[width=0.46\textwidth]{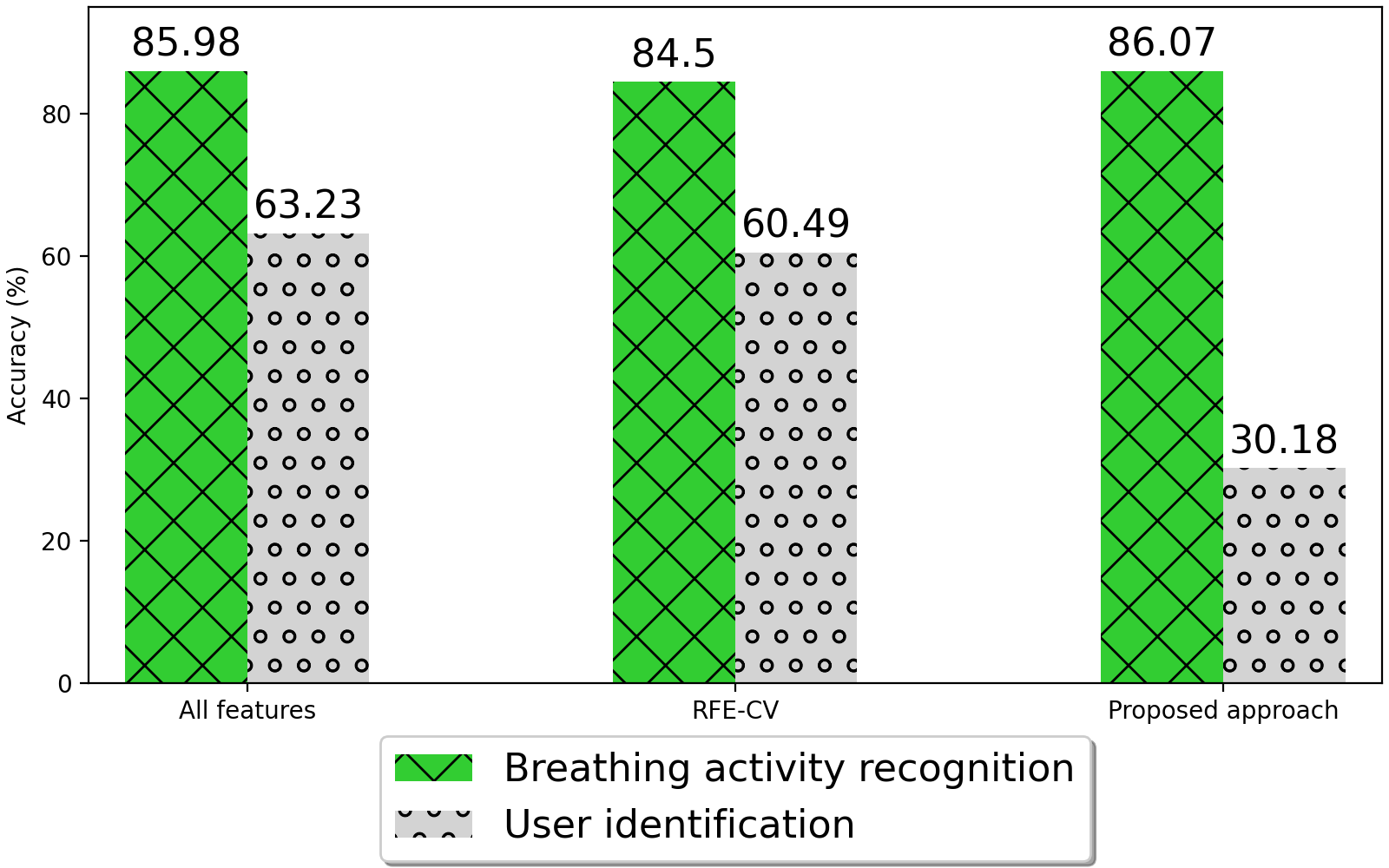}
 \end{center}
 \vspace{-3mm}
 \caption{Accuracy of classification models trained on all features, features selected by RFE with cross-validation (RFE-CV), and features selected by our proposed approach}
 \label{fig:LOGO_Comparison}
 % \vspace{-3mm}
\end{figure}

We applied a bio-inspired optimization technique~\cite{Deb2002} to select the optimal features from vital signs.
These features were used to implement two tasks: breathing activity recognition and user identification.
Our proposed approach could select feature subsets that fulfil multiple objectives: boosting the accuracy of the first task and reducing the performance of the second.
In our case, the chosen features could offer the useful task (recognizing the users' breathing patterns) while inhibiting the privacy-invading application (identifying the users).
The results of our approach can be summarized as:
\begin{itemize}
    \item Our multi-objective feature selection method increased the accuracy disparity between breathing activity recognition and user identification models up to more than 55 percentage points.
    \item We performed extensive experiments on a novel dataset of 50 subjects doing four distinct breathing activities in two positions.
    Specifically, we compared the accuracy between the full dataset and the feature-selected data in two settings: leave-one-group-out (LOGO) and leave-one-subject-out (LOSO).
    In both settings, we maintained a large disparity between the performance of breathing activity recognition and user identification.
    We applied RFE~\cite{Guyon2002} to select the optimal feature subset for breathing activity recognition and observed that these features maintained the effectiveness of user identification.
    The results are summaried in Figure~\ref{fig:LOGO_Comparison}.
    \item To reduce the computation time of the fitness function, we suggest using efficient classification models on reduced inputs.
    Our experiments show that the simplified fitness function speed up the feature selection procedure significantly (up to threefold less computation time) while conserve comparable performance disparity between these two classification models.
    \item In addition, we provide a contrariwise result which enhances the performance of user identification while reducing that of recognizing breathing patterns.
\end{itemize} 
The results show that our approach is flexible to fulfil various objectives in remote health monitoring applications.
Furthermore, using fewer features can improve the efficiency of these classification models.

\bibliographystyle{IEEEbib}
\bibliography{refs}

\end{document}